%% file: sample-sigconf.tex
\documentclass[sigconf,natbib=true,screen]{acmart}
\usepackage{xspace,balance,tabularx,multirow}
\usepackage{flushend}
\usepackage{tikz}
\usepackage{pgfplots}
\pgfplotsset{compat=1.16}
\usetikzlibrary{patterns}
\usepackage{subfig}
\usepackage[ruled, vlined, linesnumbered]{algorithm2e}
\usepackage{xcolor}
\usepackage{colortbl}
\usepackage{bbold}
\SetKwComment{Comment}{$\triangleright$\ }{}
\usepackage{enumitem}
\usepackage{tablefootnote}
\usepackage{upgreek,textgreek}
\usepackage{pifont}%
\usepackage[noabbrev]{cleveref}
\usepackage{titlecaps}
\usepackage{lipsum}

\pgfplotsset{every tick label/.append style={font=\tiny}}

\newlength{\starsize}
\newlength{\starspread}
\tikzset{starsize/.code={\setlength{\starsize}{#1}},
         starspread/.code={\setlength{\starspread}{#1}}}
\tikzset{starsize=1mm,
         starspread=3mm}
\pgfdeclarepatternformonly[\starspread,\starsize]%
  {my fivepointed stars}%
  {\pgfpointorigin}%
  {\pgfqpoint{\starspread}{\starspread}}%
  {\pgfqpoint{\starspread}{\starspread}}%
  {%
   \pgftransformshift{\pgfqpoint{\starsize}{\starsize}}
   \pgfpathmoveto{\pgfqpointpolar{18}{\starsize}}
   \pgfpathlineto{\pgfqpointpolar{162}{\starsize}}
   \pgfpathlineto{\pgfqpointpolar{306}{\starsize}}
   \pgfpathlineto{\pgfqpointpolar{90}{\starsize}}
   \pgfpathlineto{\pgfqpointpolar{234}{\starsize}}
   \pgfpathclose%
   \pgfusepath{fill}
  }

\input{tex/macro}

\AtBeginDocument{%
  \providecommand\BibTeX{{%
    \normalfont B\kern-0.5em{\scshape i\kern-0.25em b}\kern-0.8em\TeX}}}

\setcopyright{acmcopyright}
\copyrightyear{2018}
\acmYear{2018}
\acmDOI{XXXXXXX.XXXXXXX}

\acmPrice{15.00}
\acmISBN{978-1-4503-XXXX-X/18/06}

\acmSubmissionID{1787}

\begin{document}

\title{\alg{}: Cost-Effective Querying of Large Language Models for Text Clustering}
\subtitle{Technical Report}

\author{Hongtao Wang}
\affiliation{%
  \institution{Hong Kong Baptist University}
  \country{Hong Kong SAR, China}
}
\email{cshtwang@comp.hkbu.edu.hk}
\orcid{0009-0002-1279-4357}

\author{Taiyan Zhang}
\affiliation{%
  \institution{Hong Kong Baptist University}
  \country{Hong Kong SAR, China}
}
\email{zhangty2022@shanghaitech.edu.cn}
\orcid{0009-0004-6757-9237}

\author{Renchi Yang}
\authornote{Corresponding Author}
\affiliation{%
  \institution{Hong Kong Baptist University}
  \country{Hong Kong SAR, China}
}
\email{renchi@hkbu.edu.hk}
\orcid{0000-0002-7284-3096}

\author{Jianliang Xu}
\affiliation{%
  \institution{Hong Kong Baptist University}
  \country{Hong Kong SAR, China}
}
\email{xujl@comp.hkbu.edu.hk}
\orcid{0000-0001-9404-5848}

\settopmatter{printfolios=true}

\renewcommand{\shortauthors}{Wang et al.}

\begin{abstract}
Text clustering aims to automatically partition a collection of documents into coherent groups based on their linguistic features. In the literature, this task is formulated either as metric clustering over pre-trained text embeddings or as graph clustering based on pairwise similarities derived from an oracle, e.g., a large machine learning model. Recent advances in {\em large language models} (LLMs) have significantly improved this field by providing high-quality contextualized embeddings and accurate semantic similarity estimates. However, leveraging LLMs at scale introduces substantial computational and financial costs due to the large number of required API queries or inference calls.

To address this issue, we propose \alg{}, a cost-effective framework that achieves accurate text clustering under a limited budget of LLM queries. At its core, \alg{} constructs must-link and cannot-link constraints by selectively querying LLMs on informative text pairs or triplets, identified via our proposed algorithms, \algoe{} and \algot{}. These constraints are then utilized in a {\em weighted constrained clustering} algorithm to form high-quality clusters. Specifically, \algoe{} and \algot{} employ carefully designed greedy selection strategies and prompting techniques to identify and extract informative constraints efficiently. Experiments on multiple benchmark datasets demonstrate that \alg{} consistently outperforms existing methods in unsupervised text clustering under the same query budget.
\end{abstract}



\settopmatter{printfolios=true}

\maketitle

\input{tex/intro}

\input{tex/relatedwork}

\input{tex/preliminary}

\input{tex/method}

\input{tex/experiments}

\section{Conclusion}
This paper presents \alg{}, a framework that cost-effectively leverages LLMs to produce accurate pair-wise constraints for subsequent constrained text clustering. In particular, we develop two greedy algorithms \algoe{} and \algot{} to carefully select text pairs and triplets for querying LLMs so as to obtain the most informative constraints for clustering, and further ameliorate standard constrained clustering algorithms with personalized weights for various pair constraints. Our extensive experiments over benchmark datasets exhibit the high superiority of our proposed techniques in enhancing text clustering quality and reducing the sheer amount of queries needed for LLMs.

\begin{acks}
This work is partially supported by the National Natural Science Foundation of China (No. 62302414), the Hong Kong RGC ECS grant (No. 22202623), RGC grants 12202024, C1043-24GF, and C2003-23Y, and the Huawei Gift Fund.
\end{acks}

\balance
\pagebreak
\section*{GenAI Usage Disclosure}
We hereby disclose that GenAI tools were used as a component of the proposed framework in the experimental evaluation.

\bibliographystyle{ACM-Reference-Format}
\bibliography{sample-base}

\pagebreak
\appendix
\input{tex/appendix}

\end{document}

%% file: tex/macro.tex
\newcommand{\argmax}[1]{\underset{#1}{\operatorname{arg}\,\operatorname{max}}\;}
\newcommand{\argmin}[1]{\underset{#1}{\operatorname{arg}\,\operatorname{min}}\;}

\makeatletter
\newcommand*\bigcdot{\mathpalette\bigcdot@{.5}}
\newcommand*\bigcdot@[2]{\mathbin{\vcenter{\hbox{\scalebox{#2}{$\m@th#1\bullet$}}}}}
\makeatother

\newcommand{\eat}[1]{}

\newcommand{\stitle}[1]{\vspace*{0.5em}\noindent{\bf #1.\/}}

\newcommand{\G}{\mathcal{G}\xspace}

\newcommand{\T}{\mathcal{T}\xspace}

\newcommand{\Pset}{\mathcal{M}\xspace}
\newcommand{\Nset}{\mathcal{C}\xspace}

\newcommand{\Y}{\mathcal{Y}\xspace}

\newcommand{\YM}{\mathbf{Y}\xspace}

\newcommand{\xvec}{\overrightarrow{\mathbf{x}}\xspace}

\newcommand{\alg}{\texttt{Cequel}\xspace}
\newcommand{\algoe}{\texttt{EdgeLLM}\xspace}
\newcommand{\algot}{\texttt{TriangleLLM}\xspace}

\newenvironment{customlegend}[1][]{%
    \begingroup
    \csname pgfplots@init@cleared@structures\endcsname
    \pgfplotsset{#1}%
}{%
    \csname pgfplots@createlegend\endcsname
    \endgroup
}%

\def\addlegendimage{\csname pgfplots@addlegendimage\endcsname}

\makeatletter
\newcommand\footnoteref[1]{\protected@xdef\@thefnmark{\ref{#1}}\@footnotemark}
\makeatother

\let\oldnl\nl%
\newcommand{\nonl}{\renewcommand{\nl}{\let\nl\oldnl}}%

\SetKwComment{Comment}{/* }{ */}

\makeatletter 
\g@addto@macro{\@algocf@init}{\SetKwInOut{Parameter}{Parameters}} 
\makeatother

\definecolor{myred}{HTML}{fd7f6f}
\definecolor{myred_new}{HTML}{D8D8D8}
\definecolor{myred_new2}{HTML}{D7191C}
\definecolor{myblue}{HTML}{7eb0d5}
\definecolor{mygreen}{HTML}{b2e061}
\definecolor{mypurple}{HTML}{bd7ebe}
\definecolor{myorange}{HTML}{ffb55a}
\definecolor{myyellow}{HTML}{ffee65}
\definecolor{mypurple2}{HTML}{beb9db}
\definecolor{mypink}{HTML}{fdcce5}
\definecolor{mycyan}{HTML}{8bd3c7}

\definecolor{myblue2}{HTML}{115f9a}
\definecolor{myred2}{HTML}{c23728}

\definecolor{lightblue}{HTML}{90D5FF}

%% file: tex/intro.tex
\section{Introduction}
Text clustering is a fundamental task in natural language processing (NLP), which seeks to group similar text documents into clusters. This task serves as a building block bolstering a wide range of practical applications, such as query refinement~\cite{sadikov2010clustering} in search engines, user intent detection~\cite{zhang2021discovering,rodriguez2024intentgpt} in dialogue systems, social network analysis~\cite{li2024tcgnn,huang2022early}, topic discovery~\cite{yin2011geographical,castellanos2017formal} and many others.

A canonical methodology for text clustering is to encode each text into an embedding vector, followed by applying metric clustering approaches, e.g., $k$-Means, hierarchical clustering, and DBSCAN, over these text representations to generate clusters~\cite{aggarwal2012survey}. 
The effectiveness of such methods often hinges on the quality of text embeddings.
The emergence of {\em pre-trained language models} (PLMs), e.g., BERT~\cite{devlin2018bert} and its successors~\cite{liu2019roberta,reimers2019sentence,su2023one} has significantly advanced text clustering with contextual embeddings obtained via pre-training on a large corpus~\cite{muennighoff2023mteb}.
To date, {\em large language models} (LLMs) such as GPT4~\cite{openai2023gpt4}, PaLM~\cite{chowdhery2023palm}, and LLaMA~\cite{touvron2023llama} pre-trained on extensive text corpora and datasets at the expense of billions of parameters and vast computational resources achieve state-of-the-art text embeddings and have set new benchmarks in text clustering, as reported in~\cite{petukhova2024text,keraghel2024beyond}.
However, most of the powerful LLMs are deployed as paid online services, making the underlying text embeddings inaccessible or entailing computationally and financially expensive queries via APIs, particularly for a large text collection.
Moreover, due to the restricted access to LLMs, text embeddings therefrom cannot be fine-tuned and customized for specific downstream tasks, rendering the rear-mounted clustering suboptimal.

With the advent of LLMs, another way for text clustering is to obtain similarity scores of text pairs from LLMs to define a graph clustering problem~\cite{silwal2023kwikbucks}. 
This methodology enables us to model the correlations between text instances with highly accurate semantic similarities/dissimilarities from LLMs, and hence, producing high-quality clustering results~\cite{silwal2023kwikbucks,zhang2023clusterllm}.
Despite the superb performance achieved, it requires pair-wise similarities between all text instances, leading to $O(n^2)$ inference calls to the models, where $n$ is the total number of texts in the input text corpus, which is infeasible for clustering large-scale text collections in practice.
Very recently, there are a number of works~\cite{zhang2023clusterllm,wang2023goal,yang2024tec,huang2024text} exploring other ways of leveraging the feedback from instruction-tuned LLMs to guide text clustering. However, all of them still demand substantial queries to LLMs (typically with a token count of over $10\times$ the corpus size) and/or rely on skeleton support from fine-tuning embedders such as BERT, Instructor, and E5, which is time-consuming.

Motivated by this, this paper proposes \alg (short for \underline{C}ost-\underline{e}ffective \underline{que}rying of \underline{L}LMs for text clustering), which aims to maximize the text clustering quality within a fixed budget (i.e., for token consumption) for querying LLMs. To be more specific, \alg is inspired by the constrained clustering paradigm (a.k.a. semi-supervised clustering)~\cite{dinler2016survey}, which additionally takes as input a number of constraints for pairs of data points, e.g., must-links or cannot-links, from human experts.
As such, clusters can thus be obtained based on these constraints as supervision signals and cheap text embeddings from small PLMs, e.g., BERT or Sentence-BERT~\cite{reimers2019sentence}.
Given the human-level performance of LLMs in text comprehension, a simple and straightforward approach is to employ LLMs as the domain expert to create pair-wise pseudo constraints.
However, this scheme poses two pressing challenges. First, it still remains unclear which text instances should be selected for querying and how to elicit high-caliber responses from LLMs. Secondly, the answers from LLMs might be noisy or random, engendering unreliable constraints that mislead the subsequent clustering.

To tackle the above problems, \alg includes two approaches \algoe and \algot that both construct must-link and cannot-link constraints with LLMs in a cost-effective manner.
\algoe{} harnesses the prominent {\em spanning edge centrality}~\cite{teixeira2013spanning,chandra1989electrical} in graph theory to quantify the informativeness of text pairs (referred to as edges) and a well-thought-out greedy algorithm to achieve a linear-complexity identification of crucial edges. 
For each selected edge $(t_a,t_b)$, \algoe{} elicits the relation, e.g., must-link (must be in the same cluster) or cannot-link (cannot belong to the same cluster), of $t_a$ and $t_b$ from LLMs with carefully-crafted prompts containing the task context and text contents.
Based thereon, we further develop \algot that greedily selects {\em crucial} text triplets (referred to as triangles) for querying, which draws inspiration from the fact that {\em humans represent an instance through comparing with others}~\cite{nosofsky2011generalized}.
On top of that, \alg{} refines classic constrained clustering approaches, i.e., constrained spectral clustering~\cite{wang2014constrained} and constrained K-Means~\cite{wang2014constrained}, as \texttt{WCSC} and \texttt{WCKMeans}, with our carefully designed penalty weights for pairwise constraints based on the {\em pointwise mutual information}~\cite{church1990word} calculated from the text embeddings, so as to eliminate the side effects of misinformation introduced by LLMs.

Our extensive experiments comparing \alg against multiple strong baselines under the same query budget for LLMs on five benchmark text datasets demonstrate the consistent and superior performance of \alg in unsupervised text clustering tasks. In particular, on the well-known {\em Tweet} dataset, \alg is able to attain an accuracy of $73.1\%$ with a remarkable improvement of $7.29\%$ over the state-of-the-art at a monetary cost of \$0.0185 for calling GPT-4.0-Mini.

%% file: tex/relatedwork.tex
\section{Related Work}
In the sequel, we review existing studies germane to our work.

\subsection{Text Clustering}
Early methods for text clustering rely on vector space models, e.g., bag-of-words and TF-IDF~\cite{schutze2008introduction}, representing documents as {\em high-dimensional sparse} vectors. Word2Vec~\cite{mikolov2013efficient} and GloVe~\cite{pennington2014glove} introduce low-dimensional dense embeddings but fail to capture contextual variations. Transformer-based language models, e.g., BERT~\cite{devlin2018bert} and its successors~\cite{liu2019roberta,reimers2019sentence,su2023one}, mitigate these issues by generating contextual embeddings, which have set new benchmarks in text clustering.

Although embedding-based methods are simple and efficient, they can fail to capture deeper semantic nuances~\cite{sun2019fine, saha2023influence, ravi2023text, viswanathan2023large, muennighoff2023mteb}. A series of deep clustering approaches~\cite{zhou2022comprehensive} are proposed to learn representations and cluster assignments jointly. These include multi-stage~\cite{tao2021clustering, huang2014deep}, iterative~\cite{yang2016joint, caron2018deep, van2020scan, niu2022spice, chang2017deep, niu2020gatcluster}, generative~\cite{dilokthanakul2016deep}, and simultaneous methods~\cite{xie2016unsupervised, zhang2021supporting, hadifar2019self}. While effective, these approaches typically demand training on large datasets and careful tuning.

Recently, several efforts have harnessed LLMs for enhanced text clustering. Some have validated the superiority of LLM embeddings for clustering~\cite{keraghel2024beyond,petukhova2025text}. Others propose using LLMs to generate cluster labels or explanations~\cite{wang2023goal, de2023idas, huang2024text}. Methods like \texttt{ClusterLLM}~\cite{zhang2023clusterllm} use LLM-predicted relations to guide fine-tuned embedders, while others iteratively refine clusters~\cite{feng2024llmedgerefine, yang2024tec}. The active oracle process in~\cite{viswanathan2023large} is most similar to ours. Despite encouraging results, prior works strongly rely on extensive LLM queries, costly fine-tuning, and/or post-corrections. Unlike them, our method achieves superior performance with a sufficiently low number of queries and budget, without fine-tuning or post-correction.

\vspace{-2ex}
\subsection{Constrained Clustering}
To improve clustering, constrained clustering adopts side information like must-link (ML) or cannot-link (CL) constraints. Classical approaches incorporate these into traditional algorithms, such as constraint-based variants of K-means~\cite{wagstaff2001constrained} and Gaussian mixture models~\cite{shental2003computing, law2004clustering, law2005model}. A representative example is \texttt{PCKMeans}~\cite{basu2004active}, which extends K-means by adding pairwise constraints to its objective function. Spectral methods are another branch, encoding constraints in graph structures~\cite{kamvar2003spectral, kulis2005semi} or eigenspace manipulations~\cite{de2004learning}. Among these, {\em Constrained Spectral Clustering} (\texttt{CSC})~\cite{wang2014constrained} integrates constraints by modifying the graph Laplacian.

Deep learning-based approaches leverage neural networks to learn feature representations that respect constraints~\cite{hsu2015neural, chen2015deep}. Methods such as SDEC~\cite{ren2019semi} and constrained IDEC~\cite{zhang2021framework} refine latent spaces for clustering. However, generative extensions remain underexplored, with few studies~\cite{luo2018semi} integrating constraints into probabilistic frameworks. Despite progress, existing methods fail to fully exploit the structure of constraints in generative settings. To bridge this gap, we introduce weighting schemes based on \texttt{PCKMeans} and \texttt{CSC}, proposing \texttt{WCKMeans} and \texttt{WCSC}. This elevates constrained clustering from treating constraints as binary to assigning specific weights for each one.

%% file: tex/preliminary.tex
\section{Problem Formulation}

\subsection{Notation and Terminology}
Let $\T=\{t_i\}_{i=1}^n$ be an unlabeled corpus consisting of $n$ distinct text instances. 
We use $\Omega$ to denote the {\em corpus size} of $\T$, which is the total amount of text tokens in $\T$.
The text embedding of text $t_i\in \T$ is denoted as \(\xvec_i\in \mathbb{R}^d\), which can be generated via any text encoder $f(t_i)$, e.g., TF-IDF, BERT, and LLMs. We assume that each \(\xvec_i\) is $L_2$ normalized, i.e., $\|\xvec_i\|_2=1$.
Accordingly, we define the {\em embedding-based semantic similarity} (ESS) between any two text instances $t_a,t_b\in \T$ as
\begin{equation}\label{eq:cosim}
s(t_a,t_b) = \textsf{cosine}(\xvec_a, \xvec_b) = \xvec_a\cdot \xvec_b.
\end{equation}
and the total similarities of a text $t_a\in \T$ with all others as
\begin{equation}\label{eq:degree}
d(a) = \sum_{t_i\in \T}{s(t_a,t_i)} = \xvec_a \sum_{t_i\in \T}{\xvec_i}.
\end{equation}
We typically use \(\{\Y_1,\Y_2,\ldots,\Y_K\}\) to represent a set of $K$ disjoint clusters and a cluster assignment matrix $\YM\in \mathbb{R}^{n\times K}$ to symbolize the text-cluster memberships. In particular, $\YM_{i,k}=1$ if $t_i\in \Y_k$ and $0$ otherwise.
A {\em must-link} (resp. {\em cannot-link}) refers to a pair of text instances $t_i,t_j\in \T$ that are required to be in the same cluster (resp. distinct clusters).
We use \([1,n]\) to denote the set of integers \(\{1,2,\ldots,n\}\).

\subsection{Problem Statement}
Given text corpus $\T$, a query budget\footnote{The total number of tokens for consumption.} $Q$ to LLMs, and the desired (or ground-truth) number $K$ of clusters, our {\em text clustering with constraints from LLMs} (TCCL) task is to construct a {\em must-link set} (MLS) $\Pset$ and a {\em cannot-link set} (CLS) $\Nset$ based on the feedback from LLMs as supervision constraints within the query budget $Q$, such that the texts in $\T$ can be accurately partitioned into $K$ disjoint clusters \(\{\Y_1,\Y_2,\ldots,\Y_K\}\), i.e., $\Y_i\cap \Y_j=\emptyset\ \forall{i,j\in [1,K]}$.

%% file: tex/method.tex
\section{Synoptic Overview of \alg{}}\label{sec:overview}

\begin{figure}[!t]
    \centering
    \includegraphics[width=\linewidth]{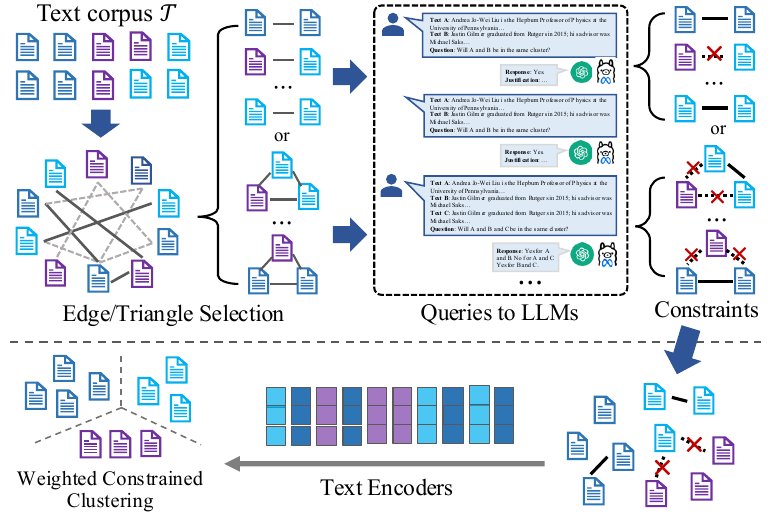}
    \vspace{-5ex}
    \caption{An overview of \alg.}
    \label{fig:overview}
    \vspace{-2ex}
\end{figure}

This section provides an overview of our proposed \alg framework for TCCL. 
As illustrated in Fig.~\ref{fig:overview}, at a high level, \alg works in a cascaded pipeline, where the first step seeks to construct MLS $\Pset$ and CLS $\Nset$, i.e., must-links or cannot-links, via our text pair/triplet queries to LLMs, while the second step produces the cluster assignments for all text instances using the {\em weighted constrained clustering} algorithms that consider the importance weight of each constraint in $\Pset$ and $\Nset$.

\stitle{Generating $\Pset$ and $\Nset$ via LLMs}
More precisely, \alg relies on \algoe or \algot to create MLS $\Pset$ and CLS $\Nset$, which build text pairs or triplets to query the LLMs to form pair-wise constraints. We refer to these two types of queries as {\em edge} and {\em triangle} queries henceforth, respectively. 
Denote by $s$ the average amount of tokens for a single text in $\T$. 
In \algoe, we first generate
\begin{equation}
N = {Q}/{(2s)}
\end{equation}
unique edges $\mathcal{S}$ from $\binom{n}{2}$ possible text pairs based on our edge centrality metric and greedy selection algorithm. 
Later, each of these edges $(t_a,t_b)\in \mathcal{S}$ will be sent to the LLM requesting an answer to whether they should belong to the same cluster or not.
Accordingly, $|\Pset|+|\Nset|=N=Q/(2s)$ must-links or cannot-links can be extracted from these ``Yes'' or ``No'' answers, forming the MLS $\Pset$ and CLS $\Nset$.

Different from \algoe, \algot selects  
\begin{equation}
N_{\triangle} = {Q}/{(3s)}
\end{equation}
distinct triangles $\mathcal{S}_\triangle$ from $\binom{n}{3}$ possible text triplets using a greedy approach. In particular, we ensure that any two triangles can have at most one common text so as to avoid duplicate pairs. \algot further obtains the relation (must-link or cannot-link) of every two texts in each $(t_a,t_b,t_c)\in \mathcal{S}_\triangle$.
Compared to edge queries, each triangle query yields three constraints. Hence, through \algot, we can generate $|\Pset|+|\Nset|=3\times N_{\triangle}=Q/s$ unique constraints in total, which is twice the amount as \algoe, when given the same query budget $Q$.

\stitle{Weighted Constrained Clustering}
Classic constrained clustering approaches, e.g., {\em Constrained Spectral Clustering} (\texttt{CSC})~\cite{wang2014constrained} and \texttt{PCKMeans}~\cite{basu2004active}, regard all must-links in $\Pset$ and cannot-links in $\Nset$ equally important. We extend them to their weighted variants, dubbed as \texttt{WCSC} and \texttt{WCKMeans}, by imposing personalized weights to various constraints based on their text embeddings. In doing so, we can offset the adverse impact of noisy must-links/cannot-links from LLMs. 

Specifically, for any constraint on text pair $(t_a,t_b)$, we leverage the prominent {\em pointwise mutual information} (PMI)~\cite{church1990word} to define the global important weight of $(t_a,t_b)$, which is formulated as follows:
\begin{small}
\begin{equation}\label{eq:PMI}
\textsf{PMI}(t_a,t_b) = \log{\left(\frac{d(a)\cdot d(b)}{s(t_a,t_b)}\cdot\sum_{t_x\in \T}{\frac{1}{d(x)}}+1\right)}.
\end{equation}
\end{small}
Given a collection $\mathcal{E}$ of elements, the PMI of an element pair $(a,b)\in \mathcal{E}$ is defined as $\textsf{PMI}(a,b)=\log(\frac{\mathbb{P}(a,b)}{\mathbb{P}(a)\cdot\mathbb{P}(b)})$, where $\mathbb{P}(a)$ (resp. $\mathbb{P}(b)$) is the probability of observing $a$ (resp. $b$) in $\mathcal{E}$ and $\mathbb{P}(a,b)$ is the probability of observing pair $(a,b)$ in $\mathcal{E}$.
The larger $\textsf{PMI}(a,b)$ is, the more likely $a$ and $b$ co-occur in $\mathcal{E}$.
In Eq.~\eqref{eq:PMI}, we let $\mathbb{P}(a)=\frac{1/d(a)}{\sum_{t_x\in \T}{\frac{1}{d(x)}}}$ (resp. $\mathbb{P}(b)=\frac{1/d(b)}{\sum_{t_x\in \T}{\frac{1}{d(x)}}}$) and $\mathbb{P}(a,b)=\frac{1/s(t_a,t_b)}{\sum_{t_x\in \T}{\frac{1}{d(x)}}}$, and add a shift value ``1'' to avoid negative output.

In \texttt{WCSC}, we first normalize the PMI values of pairs in $\Pset$ and $\Nset$ into a fixed range (e.g., \([0.5, 1.5]\)) and then construct the constraint matrix $\boldsymbol{\Theta}$ where $\boldsymbol{\Theta}_{a,b}=\textsf{PMI}(t_a,t_b)$, $-\textsf{PMI}(t_a,t_b)$ or $0$ when $(t_a,t_b)$ is a must-link, cannot-link, or has no constraints, respectively. 
$\boldsymbol{\Theta}$ is later used to build the graph Laplacian matrix for subsequent spectral clustering.

In \texttt{WCKMeans}, aside from the standard {\em within-cluster sum of squares} objective function for \texttt{K-means}, it additionally incorporates the following penalty terms for constraints in the optimization:
\begin{small}
\begin{equation}\label{eq:penality-terms}
\left( 1 + \sum_{(t_a, t_b) \in \Pset} w_{\Pset}(t_a, t_b) \right) + \left( 1 + \sum_{(t_a, t_b) \in \Nset} w_{\Nset}(t_a, t_b) \right),
\end{equation}
\end{small}
where \( w_{\Pset}(t_a, t_b) \) (resp. \( w_{\Nset}(t_a, t_b) \)) stands for the penalty weight of must-link (resp. cannot-link) \( (t_a, t_b) \).
Particularly, we adopt $\textsf{PMI}(t_a,t_b)$ as the weight but further normalize it such that \( w_{\Pset} \in [0.01, 0.1]\) and \( w_{\Nset} \in [0, 0.01]\) to balance the impact of MLS and CLS.

In the succeeding \S~\ref{sec:edge} and ~\ref{sec:triangle}, we mainly focus on elaborating on our \algoe and \algot methods for constructing MLS $\Pset$ and CLS $\Nset$. Given space limits, some details are deferred to Appendix~\ref{sec:additional_algorithms}.

\section{The \algoe Method}\label{sec:edge}
In this section, we first delineate our greedy algorithm in \algoe{} for identifying the $N$ informative edges (\S~\ref{sec:SEC} and~\ref{sec:greedy-edge}). Later, in \S~\ref{sec:edge-query-LLM}, we introduce our carefully crafted prompts to LLMs and strategies to extract must-link and cannot-link constraints.

\begin{figure}
\centering
\includegraphics[width=\linewidth]{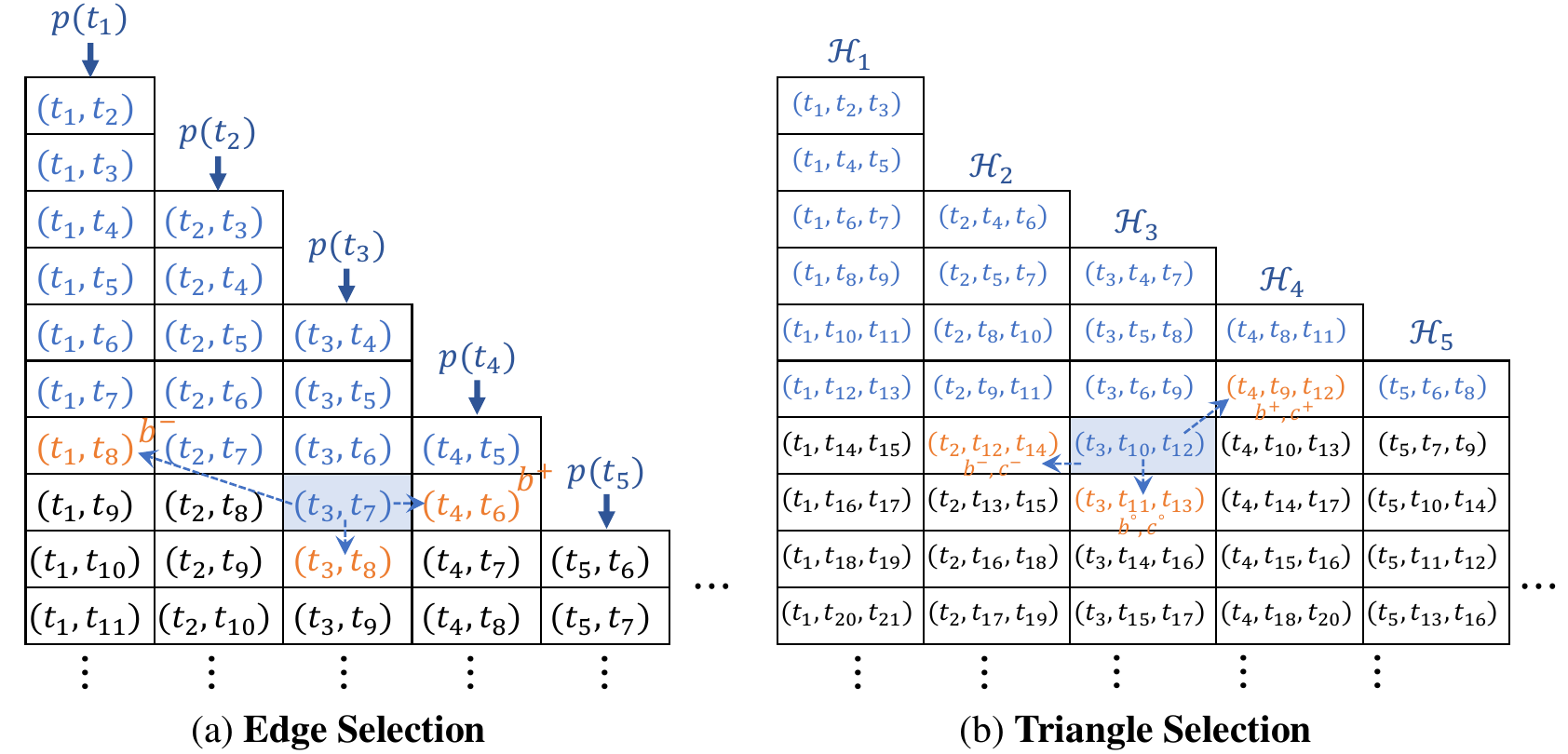}
\vspace{-4ex}
\caption{Illustration of the greedy edge/triangle selection.}
\label{fig:selection}
\vspace{-2ex}
\end{figure}

\subsection{Spanning Edge Centrality}\label{sec:SEC}
To assess the informativeness and importance of each text pair in $\T\times \T$, we propose to capitalize on the {\em spanning edge centrality} (SEC) (a.k.a. {\em effective resistance})~\cite{teixeira2013spanning,chandra1989electrical} in graph theory as the selection criteria. Firstly, we build a full affinity graph $\G$ with $\T$ as the node set and $\T\times\T$ as the edge set, wherein the weight of each edge $(t_a,t_b)$ is defined by their ESS in Eq.~\eqref{eq:cosim}. The weighted degree of any node $t_a\in \T$ is then $d(a)$ defined in Eq.~\eqref{eq:degree}.

\begin{lemma}[\cite{lai2024efficient}]\label{lem:ER}
The SEC $\Phi(t_a,t_b)$ of each edge $(t_a,t_b)\in \G$ is bounded by $\frac{1}{2} \left( \frac{1}{d(a)}+\frac{1}{d(b)} \right) \le \Phi(t_a,t_b) \le \frac{1}{1-\lambda_2} \left( \frac{1}{d(a)}+\frac{1}{d(b)} \right)$,
where $\lambda_2<1$ stands for the second largest eigenvalue of the normalized adjacency matrix of $\G$.
\end{lemma}

In an unweighted graph $\G$, the SEC $\Phi(t_a,t_b)$ of an edge $(t_a,t_b)$ can be interpreted as the fraction of spanning trees of $\G$ that contains $(t_a,t_b)$. Intuitively, a large $\Phi(t_a,t_b)$ indicates that edge $(t_a,t_b)$ plays a critical role in the graph topology.
However, the exact computation of $\Phi(t_a,t_b)$ is rather costly, and even most approximate algorithms still struggle to cope with medium-sized graphs~\cite{spielman2008graph,10.1145/3580305.3599323,yang2023efficient}.

Lemma~\ref{lem:ER} states that $\Phi(t_a,t_b)$ of any edge $(t_a,t_b)$ is proportional to the sum of the inverses of two endpoints' degrees, i.e., 
\begin{equation}
\Phi(t_a,t_b) \propto \frac{1}{d(a)} + \frac{1}{d(b)},
\end{equation}
which suggests us to use $\frac{1}{d(a)} + \frac{1}{d(b)}$ as a fast estimation of \(\Phi(t_a,t_b)\) and has been shown accurate as validated in~\cite{lai2024efficient,qiu2021lightne}. With slight abuse of notation, we let $\Phi(t_a,t_b)=\frac{1}{d(a)} + \frac{1}{d(b)}$ in the rest of this paper for tractability.

\begin{algorithm}[!t]
\caption{Greedy Edge Selection}\label{alg:edge}
\begin{small}
\KwIn{Corpus \(\T\) and the number \(N\) of edges.}
\KwOut{An edge set $\mathcal{S}$.}
Calculate the degree $t_a$ of each text $t_a\in \T$ by Eq.~\eqref{eq:degree}\;
Sort \(\T\) according to their degrees in the ascending order\;
Initialize $p(t_a)\gets a+1$ for $1\le a\le \lceil \frac{\sqrt{8N+1}-1}{2} \rceil$\;
Initialize a stack $\mathcal{S}=\{(t_1,t_2)\}$\;
Update $p(t_1)\gets 3$\;
\While{$|\mathcal{S}|< N$}{
$(t_a,t_b) \gets \mathcal{S}.\textsf{top}()$\;
${b}^{-} \gets p(t_1),\ {b}^{+} \gets p(t_{a+1})$\;
$\mathcal{B} \gets \{(t_1,t_{{b}^{-}}), (t_a,t_{b+1}), (t_{a+1},t_{{b}^{+}})\}$\;
$(t_x,t_y)\gets \argmax{(t_{a'},t_{b'})\in \mathcal{B}}{\Phi(t_{a'},t_{b'})}$\;
Push \( (t_x,t_y) \) to $\mathcal{S}$\;
$p(x)\gets y+1$\;
}
\end{small}
\end{algorithm}

\subsection{Greedy Edge Selection}\label{sec:greedy-edge}
Let \(\mathcal{A}=\{ (t_a,t_b)\in \T \times \T|t_a \neq t_b\}\). Then, the edge selection task can be formally formulated as the following maximization problem:
\begin{equation}\label{sec:obj-SEC}
\max_{\mathcal{S}\subset \mathcal{A}, |\mathcal{S}|=N} \sum_{(t_a,t_b)\in \mathcal{S}}\Phi(t_a,t_b),
\end{equation}
which seeks to extract a subset $\mathcal{S}\subseteq \mathcal{A}$ with $N$ distinct text pairs such that their total SEC over $\G$ is maximized.

A simple and straightforward way to get $\mathcal{S}$ is to calculate the SEC of each edge in $\mathcal{A}$ and pick the $N$-largest ones as $\mathcal{S}$. Such a brute-force approach entails a quadratic time complexity, i.e., $O(n^2)$, to enumerate all text pairs in $\mathcal{A}$, which is impractical for large corpora.
As a workaround, we resort to a greedy strategy for higher efficiency, whose pseudo-code is displayed in Algorithm~\ref{alg:edge}. 

\begin{lemma}\label{lem:edge-range}
$\argmax{(t_a,t_b)\in \mathcal{S},\ a<b}{a}\le \lceil \frac{\sqrt{8N+1}-1}{2} \rceil$ when $|\mathcal{S}|=N$.
\end{lemma}

Specifically, Algorithm~\ref{alg:edge} begins by reordering the text instances in \(\T\) according to their degrees defined in Eq.~\eqref{eq:degree} in the ascending order, such that $\frac{1}{d(i)}$ of the $i$-th text $t_i$ is the $i$-th largest among all texts (Lines 1-2).
Next, 
to facilitate the identification of the text $t_b$ that can yield the highest $\Phi(t_a,t_b)$ for each text $t_a$ ($b>a$), we maintain the index $p(t_a)$ of the next suggested text for each possible $t_a$, which is initialized to be $a+1$ (Line 3). According to Lemma~\ref{lem:edge-range}\footnote{The proof is deferred to Appendix~\ref{sec:lemma_proof}.}, the value of $a$ falls into the range of $[1,\lceil \frac{\sqrt{8N+1}-1}{2} \rceil]$.
At Line 4, a stack $\mathcal{S}$ is additionally created to accommodate all the selected edges and \( (t_1,t_2) \) is added to $\mathcal{S}$ as it has the globally largest SEC in $\mathcal{A}$. The index $p(t_1)$ is then increased to $3$ since $t_3$ is the text that constitutes the new best edge with $t_1$ (Line 5). At a high level, the selection is to navigate between columns for the best edges in unvisited cells (in orange) as illustrated in Fig.~\ref{fig:selection}(a).

Subsequently, Algorithm~\ref{alg:edge} starts an iterative process to greedily select the text pair $(t_a,t_b)$ ($b>a$) with the highest SEC to expand $\mathcal{S}$ (Lines 6-12). 
First, we get the latest added edge $(t_a,t_b)$ by picking the top element in stack $\mathcal{S}$ at Line 7. As such, the ideal edge $(t_x,t_y)$ to be selected in this round will be the one in $\mathcal{A}\setminus \mathcal{S}$ whose SEC is only second only to $\Phi(t_a,t_b)$. 
Simply, there are three possible choices for $(t_x,t_y)$, which satisfy $x=a$, $x>a$, or $x<a$, respectively. As per our greedy strategy, the first and second candidates are $(t_a,t_{b+1})$ and $(t_{a+1},t_{b^{+}})$, where the index $b^{+}$ can be obtained via $p(t_{a+1})$.
Intuitively, they are the two edges with the smallest SEC among all edges starting with $x=a$ and $x>a$, respectively.
As for the case with $x<a$, Algorithm~\ref{alg:edge} greedily moves to the smallest value $x=1$ and selects $(t_1,t_{b_1})$ ($b_1 = p(t_1)$) as the rest candidate (Line 8).
We then form the set $\mathcal{B}=\{(t_1,t_{{b}^{-}}), (t_a,t_{b+1}), (t_{a+1},t_{{b}^{+}})\}$ and find the one with the largest SEC as $(t_x,t_y)$, as in Lines 9-10.
Next, at Lines 11-12, we push $(t_x,t_y)$ to $\mathcal{S}$ and update the index $p(t_x)$ to $y+1$, i.e., the next best.
Algorithm~\ref{alg:edge} repeats the above procedure until the size of $\mathcal{S}$ reaches $N$ (Line 6), and returns $\mathcal{S}$ as the final result.

\stitle{Complexity Analysis}
Since $\sum_{t_i\in \T}{\xvec_i}$ in Eq.~\eqref{eq:degree} can be precomputed and reused for the degree computations of all text instances in $\T$, Line 1 in Algorithm~\ref{alg:edge} takes $O(nd)$ time. The sorting at Line 2 can be done in $O(n\log{n})$ time using the quick sort algorithm. Note that Algorithm~\ref{alg:edge} will repeat Lines 6-12 for $N$ times, and each iteration only involves $O(1)$ operations. overall, the time complexity of Algorithm~\ref{alg:edge} can thus be bounded by $O(N+n\log{n}+nd)$.

\begin{table*}[!h]
\centering
\caption{Example prompts for edge and triangle queries on CLINC.}
\vspace{-3ex}
\label{tab:unsupervised-prompt}  %
\resizebox{\textwidth}{!}{
\begin{tabular}{c|l}  %
\hline
\textbf{Type} & \textbf{Prompt} \\
\hline
\multirow{4}{*}{\centering Edge Query} & Cluster CLINC docs by whether they are the same domain category. For each pair, respond with Yes or No without explanation. \\
      & - Domain \#1: thank you ever so much for that! \\
      & - Domain \#2: i want to eat something from turkey \\
      & Given this context, do Domain \#1 and Domain \#2 likely correspond to the same domain category? \\
\hline
\multirow{7}{*}{\centering Triangle Query} & Cluster CLINC docs by whether they are the same domain category. For each triangle, respond with a, b, c, d, or e without explanation. \\
      & - Domain \#1: i am interested in a new insurance plan \\
      & - Domain \#2: go to the next song and play it \\
      & - Domain \#3: how many days of vacation do i have left \\
      & Given this context, do Domain \#1, Domain \#2 and Domain \#3 likely correspond to the same domain category? \\
      & \textbf{(a)} All are same category.\quad\quad \textbf{(b)} Only \#1 and \#2 are same category. \\
      & \textbf{(c)} Only \#1 and \#3 are same category.\quad\quad \textbf{(d)} Only \#2 and \#3 are same category.\quad\quad \textbf{(e)} None. \\
\hline
\end{tabular}
}
\end{table*}

\begin{algorithm}[!t]
\caption{\texttt{HeapAdd}}\label{alg:update}
\begin{small}
\KwIn{A min-heap \(\mathcal{H}\) and new elements $\mathcal{E}$.}
\KwOut{The updated min-heap \(\mathcal{H}\).}
$h\gets \mathcal{H}.\textsf{min}()$\;
Push elements in $\mathcal{E}$ to \(\mathcal{H}\)\;
\If{$\exists e \in \mathcal{E}\ s.t.\ e=h+1$}{
\While{$|\mathcal{H}|>1$}{
$\beta \gets \mathcal{H}.\textsf{pop}()$\;
\lIf{$\mathcal{H}.\textsf{min}()\neq \beta+1$}{
Push $\beta$ to $\mathcal{H}$ and \textbf{break}
}
}
}
\end{small}
\end{algorithm}

\begin{algorithm}[!t]
\caption{\texttt{GreedyScan}}\label{alg:scan}
\begin{small}
\KwIn{Two min-heaps \(\mathcal{H}_x\) and \(\mathcal{H}_y\).}
\KwOut{The minimum element $\gamma$.}
\lIf{$\mathcal{H}_y.\textsf{min}()+1 \notin \mathcal{H}_x$}{
$\gamma \gets \mathcal{H}_y.\textsf{min}()+1$
}\Else{
\For{$\gamma\gets \mathcal{H}_y.\textsf{min}()+2$ to $\mathcal{H}_y.\textsf{max}()+1$}{
\lIf{$\gamma\notin \mathcal{H}_y\cup \mathcal{H}_x$}{
    \textbf{break}
}
}
}
\end{small}
\end{algorithm}

\subsection{\bf Edge Queries to LLMs}\label{sec:edge-query-LLM}

Given the edge set $\mathcal{S}$ selected by Algorithm~\ref{alg:edge}, \algoe starts to query the LLM (e.g., GPT-4) for the relation of the two text instances $t_a,t_b$ in each edge $(t_a,t_b)$.
Specifically, under the unsupervised setting, \algoe first prompts the LLM with a context indicating the task being a clustering, the contents of text documents $t_a,t_b$, and the reply format: ``Yes'' or ``No'' without providing explanation. Afterwards, the LLM is instructed to answer whether $t_a$ and $t_b$ should be in the same cluster. 
If the response is ``Yes'', edge $(t_a,t_b)$ is recognized as a must-link and \algoe adds it to MLS $\Pset$, otherwise edge $(t_a,t_b)$ is appended to CLS $\Nset$.
Table~\ref{tab:unsupervised-prompt} presents the detailed prompt template for edge queries.

\section{The \algot Method}\label{sec:triangle}
This section delineates the details of \algot in forming $N_\triangle$ triangles, querying the LLM, and extracting the MLS and CLS from the responses of the LLM.

\begin{algorithm}[!t]
\caption{Greedy Triangle Selection}\label{alg:triangle}
\begin{small}
\KwIn{Corpus \(\T\) and the number \(N_\triangle\) of triangles.}
\KwOut{A triangle set $\mathcal{S}_\triangle$.}
{\nonl{Lines 1-2 are the same as Lines 1-2 in Algorithm~\ref{alg:edge}}\;
\setcounter{AlgoLine}{2}}
Initialize a min-heap $\mathcal{H}_i\gets\{i\}\ \forall{i\in [1,n]}$\;
Initialize a stack $\mathcal{S}_\triangle=\{(t_1,t_2,t_3)\}$\;
$\mathcal{H}_1, \mathcal{H}_2 \gets \texttt{HeapAdd}(\mathcal{H}_1, \{2,3\}),\ \texttt{HeapAdd}(\mathcal{H}_2, \{3\})$\;
\While{$|\mathcal{S}_\triangle|<N_\triangle$}{
$(t_a,t_b,t_c) \gets \mathcal{S}.\textsf{top}()$\;
${b}^{-}, b^{\circ}, {b}^{+}\gets \mathcal{H}_{a-1}.\textsf{min}()+1, \mathcal{H}_{a}.\textsf{min}()+1, \mathcal{H}_{a+1}.\textsf{min}()+1$\;
$c^{-}\gets \texttt{GreedyScan}(\mathcal{H}_{a-1},\mathcal{H}_{b^{-}})$\;
$c^{\circ}\gets \texttt{GreedyScan}(\mathcal{H}_{a}, \mathcal{H}_{b^{\circ}})$\;
$c^{+}\gets \texttt{GreedyScan}(\mathcal{H}_{a+1}, \mathcal{H}_{b^{+}})$\;
\(\mathcal{B} \gets \{(t_{a-1},t_{b^{-}},t_{c^{-}}),(t_{a},t_{b^{\circ}},t_{c^{\circ}}),(t_{a+1},t_{b^{+}},t_{c^{+}})\}\)\;
$(t_x,t_y,t_z)\gets \argmax{(t_{a'},t_{b'},t_{c'})\in \mathcal{B}}{\Phi_{\triangle}(t_{a'},t_{b'},t_{c'})}$\;
Push \( (t_x,t_y,t_z) \) to $\mathcal{S}_\triangle$\;
$\mathcal{H}_x,\mathcal{H}_y\gets \texttt{HeapAdd}(\mathcal{H}_x,\{y,z\}),\ \texttt{HeapAdd}(\mathcal{H}_y,\{z\})$\;
}
\end{small}
\end{algorithm}

\subsection{Spanning Triangle Centrality Maximization}
Inspired by the SEC in \S~\ref{sec:SEC}, we define the {\em spanning triangle centrality} (STC) of any triangle $(t_a,t_b,t_c)$ in $\G$ as
\begin{equation}
\Phi_{\triangle}(t_a,t_b,t_c) = \frac{\Phi(t_a,t_b)+ \Phi(t_a,t_c) + \Phi(t_b,t_c)}{2}.
\end{equation}
Recall that $\Phi(t_a,t_b)$ signifies the fraction of spanning trees in $\G$ that include edge $(t_a,t_b)$. Analogously, $\Phi_{\triangle}(t_a,t_b,t_c)$ can be roughly interpreted as the fraction of spanning trees of $\G$ where the triangle $(t_a,t_b,t_c)$ appears. Using the of trick of estimating $\Phi(\cdot,\cdot)$ in \S~\ref{sec:SEC}, we reformulate $\Phi_{\triangle}(t_a,t_b,t_c)$ as follows:  
\begin{equation}
\Phi_{\triangle}(t_a,t_b,t_c) = \frac{1}{d(a)} + \frac{1}{d(b)} + \frac{1}{d(c)}.
\end{equation}

Let \(\mathcal{A}_{\triangle}\) be the subset of \(\{ (t_a,t_b,t_c)\in \T \times \T \times \T|t_a \neq t_b\ \text{and}\ t_b\neq t_c\ \text{and}\ t_a\neq t_c\}\) such that any two distinct triangles \(\triangle_x, \triangle_y\in \mathcal{A}\) satisfy $|\triangle_x\cap \triangle_y|\le 1$. Note that this requirement ensures that no duplicate text pairs are in all selected triangles.
Similar in spirit to the objective for selecting edges in Eq.~\eqref{sec:obj-SEC}, the triangle selection is framed as the problem of selecting a subset $\mathcal{S}_{\triangle}$ of $N_{\triangle}$ triangles from \(\mathcal{A}_{\triangle}\) such that their total STC is maximized:
\begin{equation}\label{eq:obj-STC}
\max_{\mathcal{S}_{\triangle}\subseteq \mathcal{A}_{\triangle}, |\mathcal{S}_{\triangle}|=N_{\triangle}} \sum_{(t_a,t_b,t_c)\in \mathcal{S}_{\triangle}}\Phi_{\triangle}(t_a,t_b,t_c).
\end{equation}

Enumerating all possible triangles in \(\mathcal{A}_{\triangle}\) for the optimization of Eq.~\eqref{eq:obj-STC} yields a computation cost of $O(n^3)$, which is prohibitively expensive.
Instead, akin to \algoe, we greedily select the triangle $(t_a,t_b,t_c)$ ($a<b<c$) with the largest $\Phi_{\triangle}(t_a,t_b,t_c)$. To facilitate this, for each text $t_a\in \T$, we need to maintain a set $\mathcal{H}_a$ comprising text instances that occur with $t_a$ in triangles already selected. Then, to find the new triangle $(t_a,t_b,t_c)$, we can simply fetch the smallest element from $[a+1,n]\setminus \mathcal{H}_a$ as $t_b$ and then from $[b+1,n]\setminus (\mathcal{H}_a\cup \mathcal{H}_b)$ as $t_c$, respectively. Unfortunately, this approach requires materializing $\mathcal{H}_a$ consisting of many elements and sorting $[a+1,n]\setminus \mathcal{H}_a$, which is both space-consuming and computationally inefficient.

\begin{figure}[!t]
\centering
\includegraphics[width=0.99\linewidth]{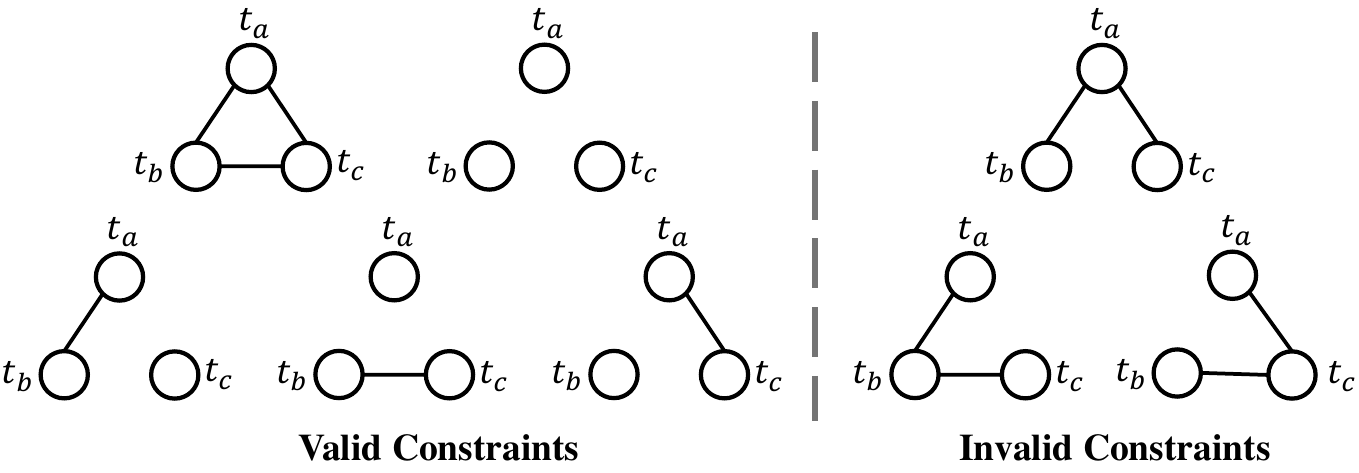}
\vspace{-2ex}
\caption{The possible results of a triangle query.}
\label{fig:triangle-result}
\vspace{-2ex}
\end{figure}

\subsection{Greedy Triangle Selection}\label{sec:triangle_selection}
To this end, we develop the {\em Greedy Triangle Selection} in Algorithm~\ref{alg:triangle}, which leverages a small-sized min-heap for each text $t_a\in \T$ to sidestep the redundant computations in searching for $t_b$ and $t_c$. As illustrated in Fig.~\ref{fig:selection}(b), akin to Algorithm~\ref{alg:edge} for edge selection, the basic idea is to greedily determine the next move over the table containing candidate triangles.

As shown in Algorithm~\ref{alg:triangle}, we first reorder texts in $\T$ as Lines 1-2 in Algorithm~\ref{alg:edge}. Then, a min-heap $\mathcal{H}_i=\{i\}$ is created for each text in $\T$ at Line 3, and a stack $\mathcal{S}_\triangle$ is initialized with adding the first triangle $(t_1,t_2,t_3)$ to optimize Eq.~\eqref{eq:obj-STC} at Line 4.
Accordingly, we update $\mathcal{H}_1$ and $\mathcal{H}_2$ using \texttt{HeapAdd} in Algorithm~\ref{alg:update} (Line 5). Basically, rather than including the indices of all text instances occurring with $t_a$ in picked triangles in $\mathcal{S}_\triangle$, our idea is to merely keep the last ``discontinuous'' index $\beta$ and those larger than $\beta$. For example, instead of maintaining $\mathcal{H}_2=\{2,3,4,6,7\}$, we refine it as $\mathcal{H}_2=\{4,6,7\}$ via \texttt{HeapAdd}.
In doing so, we can quickly identify the valid text $t_b$ for $t_a$ via $\mathcal{H}_a.\textsf{min}()+1$ for forming the triangle.
In particular, Algorithm~\ref{alg:update} iteratively pops out the minimum index $\beta$ from the min-heap $\mathcal{H}$ if the second smallest element is $\beta+1$, so as to  the last ``discontinuous'' index.

Next, Algorithm~\ref{alg:triangle} iteratively and greedily adds the new triangle to $\mathcal{S}_\triangle$ as follows. Based on the latest triangle $(t_a,t_b,t_c)$ in $\mathcal{S}_\triangle$, we examine three candidates: $(t_{a-1},t_{b^{-}},t_{c^{-}})$, $(t_{a},t_{b^{\circ}},t_{c^{\circ}})$, and $(t_{a-1},t_{b^{+}},t_{c^{+}})$ (denoted as a set $\mathcal{B}$), one of whose STCs is expected to be second only to $\Phi_{\triangle}(t_a,t_b,t_c)$ among those of $\mathcal{A}_\triangle\setminus \mathcal{S}_\triangle$. Equipped with $\mathcal{H}_{a-1}$, $\mathcal{H}_{a-1}$, and $\mathcal{H}_{a+1}$, we can easily get the best $t_{b^{-}}$, $t_{b^{\circ}}$, and $t_{b^{+}}$, respectively, as at Line 8. 

As for $t_{c^{-}}$, $t_{c^{\circ}}$, and $t_{c^{+}}$, note that ${c^{-}}$, ${c^{\circ}}$, and ${c^{+}}$ cannot appear in $\mathcal{H}_{a-1}\cup \mathcal{H}_{b^{-}}$, $\mathcal{H}_{a}\cup \mathcal{H}_{b^{\circ}}$, and $\mathcal{H}_{a+1}\cup \mathcal{H}_{b^{+}}$, and should be greater than $\mathcal{H}_{b^-}.\textsf{min}()$, $\mathcal{H}_{b^\circ}.\textsf{min}()$, and $\mathcal{H}_{b^+}.\textsf{min}()$, respectively. To implement this, we develop \texttt{GreedyScan} in Algorithm~\ref{alg:scan}, which takes input two min-heaps $\mathcal{H}_x$ and $\mathcal{H}_y$, and scans these two sets to find the minimum index $\gamma$ that is not in $\mathcal{H}_x\cup \mathcal{H}_y$ but $\gamma\ge \mathcal{H}_{y}.\textsf{min}()+1$.
At Lines 9-11, we obtain ${c^{-}}$, ${c^{\circ}}$, and ${c^{+}}$ by invoking Algorithm~\ref{alg:scan}.

Afterwards, we choose the triangle \((t_x,t_y,t_z)\) with the maximal STC from $\mathcal{B}$ (Lines 12-13), and push it to $\mathcal{S}_\triangle$ at Line 14. At Line 15, the indices $\{y,z\}$ and $\{z\}$ are then inserted into the min-heaps $\mathcal{H}_x$ and $\mathcal{H}_y$ using \texttt{HeapAdd}, respectively. The above process (Lines 7-15) is repeated until $|\mathcal{S}_\triangle|=N_\triangle$, i.e., the $N_\triangle$ triangles are found.

\stitle{Complexity Analysis}
As per our analysis in \S~\ref{sec:greedy-edge}, Lines 1-2 consume $O(n\log{n}+nd)$ time.
The time cost of each iteration at Lines 7-15 is dependent on the overhead for invoking \texttt{GreedyScan} and \texttt{HeapAdd} functions. Assume that the maximum size of the heap $\mathcal{H}_i\ \forall{i\in [1,n]}$ is $h$, then each call to \texttt{GreedyScan} is bounded by $O(h)$ as it scans all elements in the heap in the worst case. In \texttt{HeapAdd}, each insertion to $\mathcal{H}$ takes $O(\log{h})$ time, and each deletion consumes $O(1)$ time. In the worst case, \texttt{HeapAdd} will pop out $h-1$ elements. Then, each invocation of \texttt{HeapAdd} requires up to $O(h+\log{h})$ time.
Considering $N_\Delta$ iterations, the overall time complexity is therefore $O(n\log{n}+nd+N_\Delta h)$. Due to Lines 3-6, the empirical value of $h$ is rather small and thus can be regarded as a constant.

\subsection{\bf Triangle Queries to LLMs}

Subsequently, \algot requests for relations of each text triplet $(t_a,t_b,t_c)$ in $\mathcal{S}_\triangle$ returned by Algorithm~\ref{alg:triangle} from the LLM using the following prompting approach. Akin to the edge queries in \S~\ref{sec:edge-query-LLM}, the input prompts specify the task context and contents of text instances $t_a,t_b,t_c$, whereas the query is framed as a multiple-choice question (MCQ). 
In particular, the MCQ includes a set of five valid alternatives as depicted in the l.h.s. of Fig.~\ref{fig:triangle-result}, wherein a line between two text instances indicates a must-link while the empty space represents a cannot-link. 
Note that the r.h.s. three results are invalid as they lead to contradictions. For instance, when $(t_a,t_b)$ and $(t_a,t_c)$ are must-links, it can be inferred that $(t_b,t_c)$ should be a must-link, which contradicts the fact that $(t_b,t_c)$ is a cannot-link.
We instruct the LLM to respond with one of the five valid options (a)-(e) in Table~\ref{tab:unsupervised-prompt} and add the corresponding must-links and cannot-links to $\Pset$ and $\Nset$ accordingly.

%% file: tex/experiments.tex
\section{Experiments}
This section experimentally evaluates the effectiveness of the proposed \alg{} framework and its constituent techniques (\algoe{}, \algot{}, \texttt{WCSC}, and \texttt{WCKMeans}) in enhancing text clustering quality. All experiments are conducted on a Linux machine with an NVIDIA A100 GPU (80GB RAM), AMD EPYC 7513 CPU (2.6 GHz), and 1TB RAM. The source code and datasets are publicly accessible at \url{https://github.com/HKBU-LAGAS/Cequel}.

\begin{table}[!t]
\centering
\caption{Dataset statistics.}
\label{tab:stats_brief}
\vspace{-2ex}
\renewcommand{\arraystretch}{1.0}
\resizebox{\columnwidth}{!}{
\begin{small}
\begin{tabular}{c|c|c|c|c}
\hline
{\bf Corpus} & {\bf \#Texts} ($n$) & {\bf Corpus Size} ($\Omega$) & {\bf \#Clusters} ($K$) & {\bf Task} \\ \hline
BBC News & 2,225 &15,516 & 5  & News \\ 
Tweet  & 2,472 &27,883 & 89  &  Tweet\\ 
Bank77  & 3,080 &38,773 & 77  &  Intent \\ 
Reddit & 3,217 &50,969 & 50  & Topic \\ 
CLINC & 4,500 &39,970 & 10  &  Domain \\
Massive & 11,514 & 86,183 & 18  &  Scenario \\
 \hline
\end{tabular}
\end{small}
}
\label{tab:datasets}
\vspace{-2ex}
\end{table}

\subsection{Experimental Seup}
\stitle{Datasets}
Table ~\ref{tab:datasets} presents the statistics of a wide range of benchmarks for various tasks that are widely used for experimental evaluation in previous works~\cite{zhang2023clusterllm,viswanathan2023large,zhang2021supporting}. 
The corpus size $\Omega$ represents the total number of text tokens of the text instances in the respective five text corpora, which is also the token budget we want to control in the main experiments in Table \ref{tab:clustering_performance} for fair comparison.
We evaluate six benchmark datasets: {\em BBC News} (2,225 texts, 5 topics)~\cite{keraghel2024beyond}, {\em Tweet} (2,472 tweets, 89 TREC\footnote{\url{https://trec.nist.gov/data/microblog.html}} queries)~\cite{yin2016model}, {\em Bank77} (3,080 utterances, 77 intents)~\cite{zhang2021discovering}, {\em Reddit} (3,217 posts, 50 topics)~\cite{muennighoff2023mteb}, {\em CLINC} (4,500 requests, 10 domains)~\cite{zhang2022new}, and {\em Massive} (11,514 utterances, 18 scenarios)~\cite{fitzgerald2022massive}.

\stitle{Baselines}
We compare our proposed \alg{} against 10 baselines in text clustering.
First, we include 7 embedding-based approaches that adopt 7 text embedders followed by 
\texttt{K-Means++} or {\em spectral clustering}~\cite{von2007tutorial} for clustering, which include the classic \texttt{TF-IDF}, 4 popular PLMs: \texttt{E5}~\cite{wang2022text}, \texttt{DistilBERT}~\cite{reimers2019sentence}, \texttt{Sentence-BERT}~\cite{reimers2019sentence}, \texttt{Instructor-Large}~\cite{su2023one}, and 2 LLMs: \texttt{OpenAI-GPT}~\cite{brown2020language} and \texttt{LLaMA-2} (7B)~\cite{touvron2023llama}.
Moreover, we consider three recent strong methods for text clustering:  \texttt{SCCL}~\cite{zhang2021supporting} utilizes contrastive learning to enhance cluster separation; \texttt{ClusterLLM}~\cite{zhang2023clusterllm} leverages the feedback from an instruction-tuned LLM to fine-tune smaller embedding models and calibrate clustering granularity;
\texttt{PO-PCKMeans}~\cite{viswanathan2023large} obtains pairwise constraints from LLMs with a subset of ground truth examples for semi-supervised clustering.
For \texttt{ClusterLLM}, we also adopt its variant without fine-tuning for evaluation. 
The four versions of our \alg{}, i.e., \texttt{EdgeLLM} + \texttt{WCKMeans}, \texttt{EdgeLLM} + \texttt{WCSC}, \texttt{TriangleLLM} + \texttt{WCKMeans}, and \texttt{TriangleLLM} + \texttt{WCSC}, are included. 

By default, we utilize \texttt{Instructor-Large} as the primary text embedder in \texttt{SCCL}, \texttt{ClusterLLM}, \texttt{PO-PCKMeans}, and \alg{}, due to its consistently superior performance. Additionally, we employ \texttt{GPT-4o-Mini} as the backbone LLM.
For a fair comparison, we set the budget $Q$ for querying LLMs to the corpus size $\Omega$ on each dataset in \alg{}, \texttt{ClusterLLM}, \texttt{PO-PCKMeans}, and the embedding-based method \texttt{OpenAI-GPT}.
The parameters of all competitors are set as suggested in their respective papers.
Following prior studies~\cite{zhang2023clusterllm,viswanathan2023large,zhang2021supporting,feng2024llmedgerefine}, {\em clustering accuracy} (ACC) and {\em normalized mutual information} (NMI) are used as the clustering quality metrics.

\input{figs/performance}

\input{figs/ablation}

\input{figs/query-cost}

\subsection{Text Clustering Performance}
Table \ref{tab:clustering_performance} reports the ACC and NMI scores achieved by all the competitors and our \alg{} on the six datasets. 
The best results are bolded, and the best baselines are underlined.
Specifically, we can make the following observations.
First, \alg{} consistently and considerably outperforms all the baselines over all datasets, even including the semi-supervised method \texttt{PO-PCKMeans}.
For example, on {\em Tweet} and {\em CLINC} datasets, \alg{} can achieve a significant improvement of $7.29\%$ and $5.56\%$ in terms of ACC compared to the best baseline results, respectively. 
The remarkable gains validate the effectiveness of our proposed framework, which leverages greedy edge/triangle selection algorithms and LLMs to create informative pair constraints, followed by weighted constrained clustering.
Amid the four versions of \alg{}, \texttt{TriangleLLM} is always superior to \texttt{EdgeLLM} since the former yields more pair constraints compared to the latter under the same query budget $Q$, as pinpointed in \S~\ref{sec:overview}.
On {\em BBC News}, {\em Tweet}, and {\em Bank77} datasets, it can be seen that \texttt{WCSC} produces better clusters than \texttt{WCKMeans}, while the latter outperforms the former on {\em Reddit}, {\em CLINC}, and {\em Massive}.
Unless otherwise specified, we refer to \alg{} with \texttt{TriangleLLM}, \texttt{WCSC} as \alg{} for simplicity in the remaining experiments.

Another key observation we can make from Table \ref{tab:clustering_performance} is that the performance of most embedding-based approaches is subpar, except \texttt{Instructor-Large} and \texttt{OpenAI-GPT}, and \texttt{Instructor-Large} is even superior or comparable to \texttt{OpenAI-GPT} on multiple datasets, which reveals that small PLMs can also output high-quality text embeddings for clustering. However, both of them are still inferior to other LLM-based approaches, including \texttt{ClusterLLM}, \texttt{PO-PCKMeans}, and our \alg{}. The results manifest that the direct adoption of the embeddings from LLMs for text clustering is less than satisfactory, whereas using the explicit correlations between text instances from LLMs shows higher effectiveness, which is consistent with the observation in other studies~\cite{silwal2023kwikbucks,zhang2023clusterllm}.

\input{figs/ablation-selection}

\subsection{Performance when Varying Query Budgets}
In this set of experiments, we further study the text clustering performance of \alg{} and other LLM-based baselines \texttt{ClusterLLM} and \texttt{PO-PCKMeans} when the query budget $Q$ is varied. 
Fig. \ref{fig:query} depicts their ACC results on {\em BBC News}, {\em Tweet}, {\em Bank77}, and {\em Reddit} when varying $Q$ in $\{\frac{\Omega}{10},\frac{\Omega}{5},\frac{\Omega}{2},\Omega,2\Omega\}$.
It can be observed that \texttt{ClusterLLM} is unable to leverage the feedback from LLMs effectively, as its performance slightly decreases when $Q$ increases on {\em BBC News}. A similar issue can be observed in \texttt{PO-PCKMeans}, whose ACC scores undergo a remarkable decline when $Q$ is increased from $\Omega$ (resp. $\frac{\Omega}{5}$) to $2\Omega$ (resp. $\frac{\Omega}{2}$) on {\em BBC News} (resp. {\em Tweet}). In sum, both baselines fail to obtain notable improvements when expanding the budget for querying LLMs.

\input{figs/ablation-constrained-clustering}
By contrast, a consistent and conspicuous performance rise can be observed for \alg{} on all tested datasets, and the ACC scores are always considerably higher than those by \texttt{ClusterLLM} and \texttt{PO-PCKMeans} under the same query budget. In particular, when the query budget $Q=\frac{\Omega}{10}$, \alg{} achieves the clustering performance that is even superior to those attained by \texttt{ClusterLLM} and \texttt{PO-PCKMeans} with $Q=2\Omega$ on {\em BBC News} and {\em Tweet}, indicating the high query effectiveness of \alg{}.

\input{figs/ablation-constrained-weight}

\subsection{Ablation Study and Component Analysis}
This section presents ablation studies and the efficacy of each ingredient in \alg{}. Due to space constraint, we defer the results of \alg{} with various text encoders and LLMs in Appendix~\ref{sec:additional_experiments}. 

\stitle{Ablation Study}
To validate the individual contributions of our key components, we conduct a comprehensive ablation study. Specifically, we explore the effects of the proposed selection algorithms {\algoe}/{\algot} and weighted constrained clustering methods \texttt{WCKMeans}/\texttt{WCSC}. 
Taking \texttt{WCSC} and {\algot} for example, as shown in Table~\ref{tab:ablation}, the removal of \texttt{WCSC} results in a decrease in performance, especially on {\em Tweet} (ACC falls from 73.10\% to 65.33\%). This indicates that the weighting of constraints helps in managing noisy or less informative pairs. Additionally, the absence of {\algot} also causes a drop in performance, especially on {\em BBC News} (ACC declines from 91.01\% to 88.72\%), highlighting the importance of selecting high-quality constraints. These findings suggest that both components play a significant role.

\stitle{Edge/Triangle Selection Strategies}
We study three strategies in {\algot} to select triangles for querying LLMs, where the Max (resp. Min) is to maximize (resp. minimize) the SEC scores, whilst the Random generates triangles randomly. The choice of selection strategy is critical, as it directly determines the quality of constraints obtained within the limited query budget. Table \ref{tab:ablation-selection} shows that the Max strategy, as implemented by Algorithm \ref{alg:triangle}, achieves the best performance (e.g., 91.01\% ACC on BBC News). In comparison, the Min and Random strategies show significantly inferior performance due to selecting low-SEC or random triangles. The results validate the effectiveness of our selection scheme proposed in \S~\ref{sec:triangle}.

\stitle{Constrained Clustering Methods}
Once the constraints are acquired, a robust constrained clustering algorithm is essential to effectively integrate this guidance into the final partitioning process. Fig. \ref{fig:CCs-extended} presents the clustering performance of \alg{} with various constrained clustering methods, including \texttt{WCSC} (ours), \texttt{CSC}~\cite{wang2014constrained}, \texttt{PCKMeans}~\cite{basu2004active}, \texttt{COPKMeans}~\cite{bibi2023constrained}, \texttt{CCPCKMeans}~\cite{rujeerapaiboon2019size}, and \texttt{MPCKMeans}~\cite{bilenko2004integrating} on {\em BBC News}, {\em Tweet}, {\em Bank77}, and {\em Reddit}. Our \texttt{WCSC} that assigns personalized weights to constraints achieves the highest accuracy (e.g., 73.10\% ACC on {\em Tweet}). This demonstrates that \texttt{WCSC} is superior at handling the inherent noise and varying quality of LLM-generated constraints, leading to a more refined and accurate clustering outcome.

\stitle{Weighting Schemes for Constraints}
In Table~\ref{tab:clustering_results_weights}, we report the results of \alg{} when using various weighting schemes in our \texttt{WCSC} and \texttt{WCKmeans} on {\em BBC News}, {\em Tweet}, {\em Bank77}, and {\em Reddit}. 
The additional weighting schemes for comparison are listed in the following table:
\begin{table}[H]
\centering
\label{tab:weights}
\vspace{-2ex}
\renewcommand{\arraystretch}{1.0}
\begin{small}
\begin{tabular}{c|c|c}
\hline
\textbf{ESS} & \textbf{SESS}  &  \textbf{LESS} \\ \hline
$s(t_a,t_b)$ & $\sqrt{s(t_a,t_b)}$ & $\log{(s(t_a,t_b)+1)}$ \\ \hline
\textbf{SEC} & \textbf{SSEC}  &  \textbf{IPMI} \\ \hline
$\frac{1}{d(a)}+\frac{1}{d(b)}$ & $\sqrt{\frac{1}{d(a)}+\frac{1}{d(b)}}$ & $\log{\left(\frac{s(t_a,t_b)\cdot\sum_{t_x\in \T}{d(x)}}{{d(a)}\cdot {d(b)}}+1\right)}$ \\
 \hline
\end{tabular}
\end{small}
\vspace{-2ex}
\end{table}
Table~\ref{tab:clustering_results_weights} shows that improper weights for constraints lead to a significant adverse impact on the performance in both \texttt{WCSC} and \texttt{WCKmeans}. Amid all weighting schemes, the PMI defined in \S~\ref{sec:overview} consistently achieves the best results. 
For example, on {\em Tweet}, using PMI in \texttt{WCSC} improves over its unweighted version (i.e., None) by a large margin of $7.77\%$, while the best competitor IPMI can only attain an improvement of $3.56\%$.

%% file: figs/performance.tex
\begin{table*}[!t]
\centering
\renewcommand{\arraystretch}{0.95}
\caption{Clustering performance under the same query budget $Q=\Omega$. Best results are bolded, while best baselines are underlined.}
\label{tab:clustering_performance}
\vspace{-2ex}
\resizebox{\textwidth}{!}{%
\begin{tabular}{ c | c | c c | c c | c c | c c | c c | c c }
\hline
\multicolumn{2}{c|}{\textbf{Method}} 
& \multicolumn{2}{c|}{\textbf{BBC News}} 
& \multicolumn{2}{c|}{\textbf{Tweet}} 
& \multicolumn{2}{c|}{\textbf{Bank77}} 
& \multicolumn{2}{c|}{\textbf{Reddit}} 
& \multicolumn{2}{c|}{\textbf{CLINC}} 
& \multicolumn{2}{c}{\textbf{Massive}} \\
\cline{3-14}
\textbf{Embedding} & \textbf{Clustering} & ACC & NMI & ACC & NMI & ACC & NMI & ACC & NMI & ACC & NMI & ACC & NMI \\
\hline
\multirow{2}{*}{\texttt{TF-IDF}} 
& \texttt{K-means++} & 24.88 & 1.84 & 55.19 & 77.62 & 36.35 & 58.83 & 12.87 & 18.66 & 25.38 & 14.09 & 32.03 & 30.00 \\
& \texttt{SpectralClust} & 23.90 & 1.38 & 48.22 & 67.26 & 33.19 & 53.58 & 10.70 & 16.73 & 28.64 & 16.45 & 34.15 & 29.35 \\
\hline
\multirow{2}{*}{\texttt{E5}} 
& \texttt{K-means++} & 71.66 & 53.03 & 60.90 & 85.91 & 58.74 & 77.09 & 38.88 & 48.75 & \underline{59.60} & 56.25 & 41.56 & 42.36 \\
& \texttt{SpectralClust} & 63.27 & 43.87 & 59.33 & 84.75 & 57.18 & 76.20 & 45.74 & 53.06 & 53.49 & 48.87 & 46.05 & 46.76 \\
\hline
\multirow{2}{*}{\texttt{DistilBERT}} 
& \texttt{K-means++} & 74.74 & 49.03 & 48.35 & 76.47 & 31.12 & 53.38 & 26.14 & 35.24 & 41.50 & 34.23 & 30.04 & 28.13 \\
& \texttt{SpectralClust} & 71.42 & 41.82 & 51.08 & 77.62 & 39.71 & 60.45 & 30.40 & 38.44 & 35.47 & 27.79 & 29.23 & 29.51 \\
\hline
\multirow{2}{*}{\texttt{Sentence-BERT}} 
& \texttt{K-means++} & 79.02 & 56.88 & 62.74 & 86.42 & 60.77 & 78.84 & 42.31 & 50.02 & 50.32 & 48.98 & 59.21 & 64.17 \\
& \texttt{SpectralClust} & 59.55 & 42.49 & 58.30 & 83.84 & 58.52 & 76.58 & 43.29 & 50.18 & 44.24 & 45.74 & 54.74 & 60.69 \\
\hline
\multirow{2}{*}{\texttt{Instructor-Large}} 
& \texttt{K-means++} & 86.53 & 71.51 & 65.34 & 87.74 & 64.57 & 82.01 & 53.62 & 62.62 & 55.94 & 57.41 & 59.19 & 66.21 \\
& \texttt{SpectralClust} & 84.45 & 62.43 & 63.14 & 86.85 & 60.76 & 79.03 & \underline{56.64} & \underline{62.89} & 52.64 & 55.49 & 58.53 & 60.63 \\
\hline
\multirow{2}{*}{\texttt{OpenAI-GPT}} 
& \texttt{K-means++} & 88.56 & \underline{73.65} & 65.42 & 87.64 & 64.20 & 81.74 & 51.06 & 60.10 & 55.31 & 56.05 & 62.28 & \underline{68.08} \\
& \texttt{SpectralClust} & 67.01 & 51.38 & 63.00 & 86.20 & 58.90 & 77.95 & 54.23 & 61.10 & 58.04 & 53.28 & 60.76 & 63.16 \\
\hline
\multirow{2}{*}{\texttt{LLaMA-2} (7B)} 
& \texttt{K-means++} & 59.71 & 30.60 & 40.93 & 69.58 & 17.86 & 38.92 & 17.82 & 27.46 & 25.01 & 13.87 & 18.92 & 15.17 \\
& \texttt{SpectralClust} & 49.05 & 24.11 & 52.29 & 76.27 & 27.60 & 49.33 & 24.43 & 34.67 & 27.08 & 18.46 & 25.61 & 21.20 \\
\hline
\multicolumn{2}{c|}{\texttt{SCCL}~\cite{zhang2021supporting}} & 83.60 & 62.00 & 36.50 & 68.50 & 36.00 & 61.30 & 28.70 & 36.00 & 46.60 & 44.00 & 45.90 & 45.93 \\
\hline
\multicolumn{2}{c|}{\texttt{ClusterLLM}~\cite{zhang2023clusterllm}} & 88.40 & 69.73 & \underline{65.81} & \underline{88.61} & \underline{68.27} & \underline{82.45} & 54.80 & 62.47 & 51.90 & 55.04 & 61.02 & 67.92 \\
\multicolumn{2}{c|}{\texttt{ClusterLLM} w/o finetune~\cite{zhang2023clusterllm}} & 89.07 & 70.97 & 64.59 & 87.78 & 65.14 & 82.33 & 54.75 & 62.87 & {58.66} & \underline{58.95} & 59.37 & 66.60 \\
\hline
\multicolumn{2}{c|}{\texttt{PO-PCKMeans}~\cite{viswanathan2023large}} & \underline{89.53} & 72.16 & 65.58 & 88.20 & 67.79 & 81.92 & 55.10 & 62.51 & 57.80 & 56.32 & \underline{64.11} & 65.99 \\
\hline
\multicolumn{2}{c|}{\alg{} (\algoe + \texttt{WCKMeans})} & 90.47 & 73.96 & 71.16 & 88.90 & 65.29 & 82.52 & 57.07 & 63.80 & 62.31 & 61.11 & 67.66 & 68.16 \\
\multicolumn{2}{c|}{\alg{} (\algoe + \texttt{WCSC})} & 86.88 & 69.49 & 70.91 & 88.51 & 67.99 & 82.55 & 57.01 & 64.25 & 56.20 & 57.77 & 60.50 & 66.05 \\
\multicolumn{2}{c|}{\alg{} (\algot + \texttt{WCKMeans})} & 90.65 & 74.25 & 72.25 & 88.60 & 66.72 & 83.07 & \textbf{59.28} & \textbf{64.98} & \textbf{65.16} & \textbf{62.09} & \textbf{69.13} & \textbf{70.44} \\
\multicolumn{2}{c|}{\alg{} (\algot + \texttt{WCSC})} & \textbf{91.01} & \textbf{75.22} & \textbf{73.10} & \textbf{89.08} & \textbf{69.03} & \textbf{83.20} & 57.78 & 64.25 & 57.91 & 60.02 & 61.51 & 67.05 \\
\hline
\end{tabular}
}
\end{table*}

%% file: figs/ablation.tex
\begin{table}[!t]
\centering
\renewcommand{\arraystretch}{0.9}
\caption{Ablation study on \alg{}.}
\label{tab:ablation}
\vspace{-2ex}
\resizebox{\columnwidth}{!}{%
\begin{tabular}{c|cc|cc|cc|cc}
\toprule
\multirow{2}{*}{\textbf{Variant}} 
& \multicolumn{2}{c|}{\textbf{BBC News}} 
& \multicolumn{2}{c|}{\textbf{Tweet}} 
& \multicolumn{2}{c|}{\textbf{Bank77}} 
& \multicolumn{2}{c}{\textbf{Reddit}} \\
\cline{2-9}
& ACC & NMI 
& ACC & NMI 
& ACC & NMI 
& ACC & NMI \\
\midrule
\alg{}  
& \textbf{90.47} & \textbf{73.96} & \textbf{71.16} & \textbf{88.90} & \textbf{65.29} & \textbf{82.52} & \textbf{57.07} & \textbf{63.80} \\
w/o \texttt{WCKMeans} & 88.65 & 71.82 & 69.30 & 87.95 & 63.25 & 81.25 & 55.48 & 62.15 \\
w/o \algoe             & 89.12 & 72.35 & 68.85 & 87.67 & 64.13 & 81.85 & 56.12 & 62.97 \\
\midrule
\alg{}
& \textbf{86.88} & \textbf{69.49} & \textbf{70.91} & \textbf{88.51} & \textbf{67.99} & \textbf{82.55} & \textbf{57.01} & \textbf{64.25} \\
w/o \texttt{WCSC} & 85.21 & 67.94 & 68.25 & 87.68 & 66.44 & 81.71 & 55.52 & 63.24 \\
w/o \algoe        & 85.80 & 68.55 & 67.91 & 87.48 & 66.85 & 81.95 & 56.32 & 63.45 \\
\midrule
\alg{}  
& \textbf{90.65} & \textbf{74.25} & \textbf{72.25} & \textbf{88.60} & \textbf{66.72} & \textbf{83.07} & \textbf{59.28} & \textbf{64.98} \\
w/o \texttt{WCKMeans} & 89.80 & 72.38       & 65.66 & 87.84 & 64.85 & 82.17 & 57.22 & 64.12 \\
w/o \algot             & 88.25 & 71.60 & 69.53 & 87.81 & 65.16 & 81.75 & 56.57 & 63.59 \\
\midrule
\alg{} 
& \textbf{91.01} & \textbf{75.22} & \textbf{73.10} & \textbf{89.08} & \textbf{69.03} & \textbf{83.20} & \textbf{57.78} & \textbf{64.25} \\
w/o \texttt{WCSC} & 90.43 & 73.90 & 65.33 & 87.84 & 67.12 & 82.65 & 56.30 & 63.74 \\
w/o \algot & 88.72 & 70.25 & 67.27 & 87.44 & 66.49 & 83.10 & 57.22 & 63.83 \\
\bottomrule
\end{tabular}
}
\vspace{-2ex}
\end{table}

%% file: figs/query-cost.tex
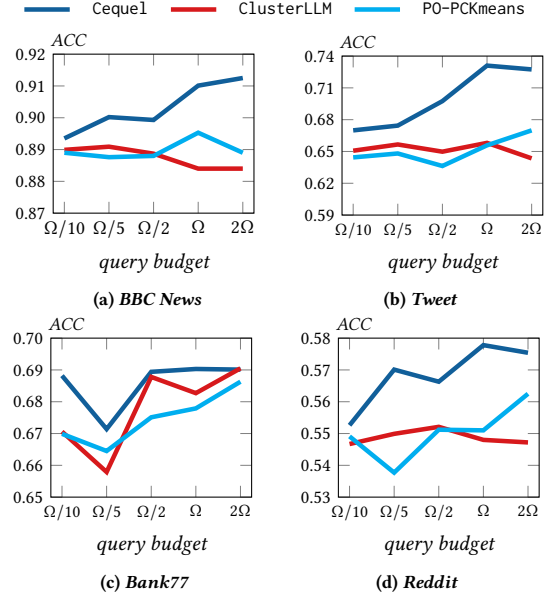
\begin{figure}[!t]
\centering
\begin{small}
\begin{tikzpicture}
    \begin{customlegend}
    [legend columns=3,
        legend entries={\alg{}, \texttt{ClusterLLM}, \texttt{PO-PCKmeans}},
        legend style={at={(0.45,1.35)},anchor=north,draw=none,font=\footnotesize,column sep=0.2cm}]
    \addlegendimage{line width=0.6mm,mark size=4pt,mark=none,color=myblue2}
    \addlegendimage{line width=0.6mm,mark size=4pt,mark=none,color=myred_new2}
    \addlegendimage{line width=0.6mm,mark size=4pt,mark=none,color=cyan}
    \end{customlegend}
\end{tikzpicture}
\\[-\lineskip]
\vspace{-3ex}
\subfloat[\em BBC News]{
\begin{tikzpicture}[scale=1,every mark/.append style={mark size=2pt}]
    \begin{axis}[
        height=\columnwidth/2.3,
        width=\columnwidth/2.0,
        ylabel={\it ACC},
        xlabel={\em query budget},
        xmin=0.5, xmax=9.5,
        xtick={1,3,5,7,9},
        xticklabel style = {font=\footnotesize},
        yticklabel style = {font=\footnotesize},
        xticklabels={${\Omega}/{10}$,${\Omega}/{5}$,${\Omega}/{2}$,$\Omega$,$2\Omega$},
        ymin=0.87,
        ymax=0.92,
        ytick={0.87,0.88,0.89,0.90,0.91,0.92},
        yticklabels={0.87,0.88,0.89,0.90,0.91,0.92},
        every axis y label/.style={font=\footnotesize,at={(current axis.north west)},right=2mm,above=0mm},
        legend style={fill=none,font=\small,at={(0.02,0.99)},anchor=north west,draw=none},
    ]
    \addplot[line width=0.6mm, mark=none,color=myblue2]  %
        plot coordinates {
(1,	0.8935	)
(3,	0.9002	)
(5,	0.8993	)
(7,	0.9101	)
(9,	0.9125	)
    };

   \addplot[line width=0.6mm, mark=none,color=myred_new2]  %
        plot coordinates {
(1,	0.8899 )
(3,	0.8909	)
(5,	0.8887	)
(7,	0.8840	)
(9,	0.8840	)
    };

   \addplot[line width=0.6mm, mark=none,color=cyan]  %
        plot coordinates {
(1, 0.8890	)
(3, 0.8876	)
(5, 0.8880	)
(7, 0.8953	)
(9, 0.8890	)
    };

    \end{axis}
\end{tikzpicture}\hspace{4mm}\label{fig:alpha-pubmed}%
}
\subfloat[\em Tweet]{
\begin{tikzpicture}[scale=1,every mark/.append style={mark size=2pt}]
    \begin{axis}[
        height=\columnwidth/2.3,
        width=\columnwidth/2.0,
        ylabel={\it ACC},
        xlabel={\em query budget},
        xmin=0.5, xmax=9.5,
        xtick={1,3,5,7,9},
        xticklabel style = {font=\scriptsize},
        yticklabel style = {font=\footnotesize},
        xticklabels={${\Omega}/{10}$,${\Omega}/{5}$,${\Omega}/{2}$,$\Omega$,$2\Omega$},
        ymin=0.59,
        ymax=0.74,
        ytick={0.59,0.62,0.65,0.68,0.71,0.74},
        yticklabels={0.59,0.62,0.65,0.68,0.71,0.74},
        every axis y label/.style={font=\footnotesize,at={(current axis.north west)},right=2mm,above=0mm},
        legend style={fill=none,font=\small,at={(0.02,0.99)},anchor=north west,draw=none},
    ]
    \addplot[line width=0.6mm, mark=none,color=myblue2]  %
        plot coordinates {
(1,	0.6699	)
(3,	0.6744	)
(5,	0.6975	)
(7,	0.7310	)
(9,	0.7274	)
    };

   \addplot[line width=0.6mm, mark=none,color=myred_new2]  %
        plot coordinates {
(1,	0.6507 )
(3,	0.6566	)
(5,	0.6498	)
(7,	0.6581	)
(9,	0.6435	)
    };

   \addplot[line width=0.6mm, mark=none,color=cyan]  %
        plot coordinates {
(1, 0.6444	)
(3, 0.6481	)
(5, 0.6363	)
(7, 0.6558	)
(9, 0.6699	)
    };

    \end{axis}
\end{tikzpicture}\hspace{0mm}\label{fig:alpha-dblp}%
}
\\[-\lineskip]
\vspace{-3ex}
\subfloat[\em Bank77]{
\begin{tikzpicture}[scale=1,every mark/.append style={mark size=2pt}]
    \begin{axis}[
        height=\columnwidth/2.3,
        width=\columnwidth/2.0,
        ylabel={\it ACC},
        xlabel={\em query budget},
        xmin=0.5, xmax=9.5,
        xtick={1,3,5,7,9},
        xticklabel style = {font=\scriptsize},
        yticklabel style = {font=\footnotesize},
        xticklabels={${\Omega}/{10}$,${\Omega}/{5}$,${\Omega}/{2}$,$\Omega$,$2\Omega$},
        ymin=0.65,
        ymax=0.70,
        ytick={0.65,0.66,0.67,0.68,0.69,0.70},
        yticklabels={0.65,0.66,0.67,0.68,0.69,0.70},
        every axis y label/.style={font=\footnotesize,at={(current axis.north west)},right=2mm,above=0mm},
        legend style={fill=none,font=\small,at={(0.02,0.99)},anchor=north west,draw=none},
    ]
    \addplot[line width=0.6mm, mark=none,color=myblue2]  %
        plot coordinates {
(1,	0.6882	)
(3,	0.6714	)
(5,	0.6894	)
(7,	0.6903	)
(9,	0.6901	)
    };

   \addplot[line width=0.6mm, mark=none,color=myred_new2]  %
        plot coordinates {
(1,	0.6706 )
(3,	0.6579	)
(5,	0.6878	)
(7,	0.6827	)
(9,	0.6905	)
    };

   \addplot[line width=0.6mm, mark=none,color=cyan]  %
        plot coordinates {
(1, 0.6699	)
(3, 0.6645	)
(5, 0.6751	)
(7, 0.6779	)
(9, 0.6863	)
    };

    \end{axis}
\end{tikzpicture}\hspace{4mm}\label{fig:alpha-dblp}%
}
\subfloat[\em Reddit]{
\begin{tikzpicture}[scale=1,every mark/.append style={mark size=2pt}]
    \begin{axis}[
        height=\columnwidth/2.3,
        width=\columnwidth/2.0,
        ylabel={\it ACC},
        xlabel={\em query budget},
        xmin=0.5, xmax=9.5,
        xtick={1,3,5,7,9},
        xticklabel style = {font=\scriptsize},
        yticklabel style = {font=\footnotesize},
        xticklabels={${\Omega}/{10}$,${\Omega}/{5}$,${\Omega}/{2}$,$\Omega$,$2\Omega$},
        ymin=0.53,
        ymax=0.58,
        ytick={0.53,0.54,0.55,0.56,0.57,0.58,0.59},
        yticklabels={0.53,0.54,0.55,0.56,0.57,0.58,0.59},
        every axis y label/.style={font=\footnotesize,at={(current axis.north west)},right=2mm,above=0mm},
        legend style={fill=none,font=\small,at={(0.02,0.99)},anchor=north west,draw=none},
    ]
    \addplot[line width=0.6mm, mark=none,color=myblue2]  %
        plot coordinates {
(1,	0.5526	)
(3,	0.5701	)
(5,	0.5663	)
(7,	0.5778	)
(9,	0.5754	)
    };

   \addplot[line width=0.6mm, mark=none,color=myred_new2]  %
        plot coordinates {
(1,	0.5467 )
(3,	0.5499	)
(5,	0.5521	)
(7,	0.5480	)
(9,	0.5472	)
    };

   \addplot[line width=0.6mm, mark=none,color=cyan]  %
        plot coordinates {
(1, 0.5491	)
(3, 0.5377	)
(5, 0.5512	)
(7, 0.5510	)
(9, 0.5625	)
    };

    \end{axis}
\end{tikzpicture}\hspace{0mm}\label{fig:alpha-dblp}%
}
\end{small}
\vspace{-3ex}
\caption{Clustering performance when varying $Q$} \label{fig:query}
\vspace{-2ex}
\end{figure}

%% file: figs/ablation-selection.tex
\begin{table}[!t]
\centering
\renewcommand{\arraystretch}{0.9}
\caption{\alg{} with various selection strategies.}
\label{tab:ablation-selection}
\vspace{-2ex}
\begin{tabular}{llcc}
\toprule
\textbf{Dataset} & \textbf{Strategy} & \textbf{ACC} & \textbf{NMI} \\
\midrule
\multirow{3}{*}{BBC News}    
       & Max (Ours)        & \textbf{91.01} & \textbf{75.22} \\
       & Min               & 67.42 & 64.70 \\
       & Random            & 88.72 & 70.25 \\
\hline
\multirow{3}{*}{Tweet}  
       & Max (Ours)        & \textbf{73.10} & \textbf{89.08} \\
       & Min               & 66.63 & 87.12 \\
       & Random            & 67.27 & 87.44 \\
\hline
\multirow{3}{*}{Bank77}  
       & Max (Ours)        & \textbf{69.03} & \textbf{83.20} \\
       & Min               & 65.84 & 82.58 \\
       & Random            & 66.49 & 83.10 \\
\hline
\multirow{3}{*}{Reddit}  
       & Max (Ours)        & \textbf{57.78} & \textbf{64.25} \\
       & Min               & 57.36 & 64.03 \\
       & Random            & 57.22 & 63.83 \\
\bottomrule
\end{tabular}
\vspace{-2ex}
\end{table}

%% file: figs/ablation-constrained-clustering.tex
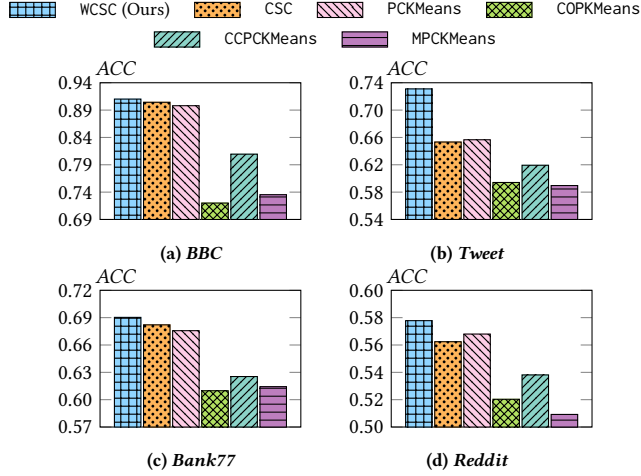
\begin{figure}[!t]
\centering
\begin{small}
\begin{tikzpicture}
\begin{customlegend}[
        legend entries={\texttt{WCSC} (Ours),\texttt{CSC},\texttt{PCKMeans},\texttt{COPKMeans}},
        legend columns=4,
        area legend,
        legend style={at={(0.45,1.15)},anchor=north,draw=none,font=\footnotesize,column sep=0.25cm}]
        \addlegendimage{preaction={fill, lightblue}, pattern={grid}}   
        \addlegendimage{preaction={fill, myorange}, pattern={crosshatch dots}} 
        \addlegendimage{preaction={fill, mypink},pattern=north west lines} 
        \addlegendimage{preaction={fill, mygreen},pattern=crosshatch}    
    \end{customlegend}
    \begin{customlegend}[
        legend entries={\texttt{CCPCKMeans},\texttt{MPCKMeans}},%
        legend columns=3,
        area legend,
        legend style={at={(0.45,0.75)},anchor=north,draw=none,font=\footnotesize,column sep=0.25cm}]  
        \addlegendimage{preaction={fill, mycyan},pattern=north east lines}    
        \addlegendimage{preaction={fill, mypurple},pattern=horizontal lines}    
    \end{customlegend}
\end{tikzpicture}
\\[-\lineskip]
\vspace{-4mm}

\subfloat[{\em BBC}]{
\begin{tikzpicture}[scale=1]
\begin{axis}[
    height=\columnwidth/2.5,
    width=\columnwidth/2.0,
    xtick=\empty,
    ybar=1.0pt,
    bar width=0.35cm,
    enlarge x limits=true,
    ylabel={\em ACC},
    ymin=0.69,
    ymax=0.94,
    ytick={0.69,0.74,0.79,0.84,0.89,0.94},
    yticklabels={0.69,0.74,0.79,0.84,0.89,0.94},
    xticklabel style = {font=\small},
    yticklabel style = {font=\small},
    every axis y label/.style={at={(current axis.north west)},right=2mm,above=0mm},
    ]

\addplot [preaction={fill, lightblue}, pattern={grid}] coordinates {(1,0.9101) }; 
\addplot [preaction={fill, myorange}, pattern={crosshatch dots}] coordinates {(1,0.9043) }; 
\addplot [preaction={fill, mypink},pattern=north west lines] coordinates {(1,0.8980) }; 
\addplot [preaction={fill, mygreen},pattern=crosshatch] coordinates {(1,0.7200) };  
\addplot [preaction={fill, mycyan},pattern=north east lines] coordinates {(1,0.8094) }; 
\addplot [preaction={fill, mypurple},pattern=horizontal lines] coordinates {(1,0.7353) };
\end{axis}
\end{tikzpicture}\hspace{4mm}
}%
\subfloat[{\em Tweet}]{
\begin{tikzpicture}[scale=1]
\begin{axis}[
    height=\columnwidth/2.5,
    width=\columnwidth/2.0,
    xtick=\empty,
    ybar=1.0pt,
    bar width=0.35cm,
    enlarge x limits=true,
    ylabel={\em ACC},
    ymin=0.54,
    ymax=0.74,
    ytick={0.54,0.58,0.62,0.66,0.70,0.74},
    yticklabels={0.54,0.58,0.62,0.66,0.70,0.74},
    xticklabel style = {font=\small},
    yticklabel style = {font=\small},
    every axis y label/.style={at={(current axis.north west)},right=2mm,above=0mm},
    ]

\addplot [preaction={fill, lightblue}, pattern={grid}] coordinates {(1,0.7310) }; 
\addplot [preaction={fill, myorange}, pattern={crosshatch dots}] coordinates {(1,0.6533) }; 
\addplot [preaction={fill, mypink},pattern=north west lines] coordinates {(1,0.6566) }; 
\addplot [preaction={fill, mygreen},pattern=crosshatch] coordinates {(1,0.5941) };  
\addplot [preaction={fill, mycyan},pattern=north east lines] coordinates {(1,0.6193) }; 
\addplot [preaction={fill, mypurple},pattern=horizontal lines] coordinates {(1,0.5896) };
\end{axis}
\end{tikzpicture}\hspace{0mm}
}%
\\[-\lineskip]
\vspace{-4mm}

\subfloat[{\em Bank77}]{
\begin{tikzpicture}[scale=1]
\begin{axis}[
    height=\columnwidth/2.5,
    width=\columnwidth/2.0,
    xtick=\empty,
    ybar=1.0pt,
    bar width=0.35cm,
    enlarge x limits=true,
    ylabel={\em ACC},
    ymin=0.57,
    ymax=0.72,
    ytick={0.57,0.60,0.63,0.66,0.69,0.72},
    yticklabels={0.57,0.60,0.63,0.66,0.69,0.72},
    xticklabel style = {font=\small},
    yticklabel style = {font=\small},
    every axis y label/.style={at={(current axis.north west)},right=2mm,above=0mm},
    ]

\addplot [preaction={fill, lightblue}, pattern={grid}] coordinates {(1,0.6903) }; 
\addplot [preaction={fill, myorange}, pattern={crosshatch dots}] coordinates {(1,0.6821) }; 
\addplot [preaction={fill, mypink},pattern=north west lines] coordinates {(1,0.6757) }; 
\addplot [preaction={fill, mygreen},pattern=crosshatch] coordinates {(1,0.6098) };  
\addplot [preaction={fill, mycyan},pattern=north east lines] coordinates {(1,0.6254) }; 
\addplot [preaction={fill, mypurple},pattern=horizontal lines] coordinates {(1,0.6143) };
\end{axis}
\end{tikzpicture}\hspace{4mm}
}%
\subfloat[{\em Reddit}]{
\begin{tikzpicture}[scale=1]
\begin{axis}[
    height=\columnwidth/2.5,
    width=\columnwidth/2.0,
    xtick=\empty,
    ybar=1.0pt,
    bar width=0.35cm,
    enlarge x limits=true,
    ylabel={\em ACC},
    ymin=0.50,
    ymax=0.60,
    ytick={0.50,0.52,0.54,0.56,0.58,0.60},
    yticklabels={0.50,0.52,0.54,0.56,0.58,0.60},
    xticklabel style = {font=\small},
    yticklabel style = {font=\small},
    every axis y label/.style={at={(current axis.north west)},right=2mm,above=0mm},
    ]

\addplot [preaction={fill, lightblue}, pattern={grid}] coordinates {(1,0.5778) }; 
\addplot [preaction={fill, myorange}, pattern={crosshatch dots}] coordinates {(1,0.5624) }; 
\addplot [preaction={fill, mypink},pattern=north west lines] coordinates {(1,0.5680) }; 
\addplot [preaction={fill, mygreen},pattern=crosshatch] coordinates {(1,0.5203) };  
\addplot [preaction={fill, mycyan},pattern=north east lines] coordinates {(1,0.5381) }; 
\addplot [preaction={fill, mypurple},pattern=horizontal lines] coordinates {(1,0.5092) };
\end{axis}
\end{tikzpicture}\hspace{0mm}
}%

\end{small}
\vspace{-2ex}
\caption{Varying constrained clustering methods in \alg{}.} 
\label{fig:CCs-extended}
\vspace{-1ex}
\end{figure}

%% file: figs/ablation-constrained-weight.tex
\begin{table}[!t]
\centering
\begin{small}
\caption{Various weighting schemes for constraints.}
\label{tab:clustering_results_weights}
\vspace{-2ex}
\renewcommand{\arraystretch}{0.85}
\resizebox{\columnwidth}{!}{%
\begin{tabular}{c|c|cc|cc|cc|cc}
\toprule
\multirow{2}{*}{} & \multirow{2}{*}{\textbf{Weight}} 
& \multicolumn{2}{c|}{\textbf{BBC News}} 
& \multicolumn{2}{c|}{\textbf{Tweet}} 
& \multicolumn{2}{c|}{\textbf{Bank77}} 
& \multicolumn{2}{c}{\textbf{Reddit}} \\
\cline{3-10}
& & \textbf{ACC} & \textbf{NMI} 
& \textbf{ACC} & \textbf{NMI} 
& \textbf{ACC} & \textbf{NMI} 
& \textbf{ACC} & \textbf{NMI} \\
\midrule
\multirow{8}{*}{\rotatebox[origin=c]{90}{\texttt{WCSC}}} 	
& None     & \underline{90.43} & \underline{73.90} & 65.33 & 87.84 & 67.12 & 82.65 & 56.30 & 63.74 \\
& ESS      & 72.40  & 65.31      & 66.55 & 87.22  & 66.84 & 81.29 & 55.41 & 62.95 \\
& SESS     & 71.37  & 63.35      & 66.34 & 87.17  & 67.45 & 81.75 & 56.08 & 62.74 \\
& LESS     & 71.15  & 64.36      & 65.33 & 86.85  & 67.32 & 81.61 & 54.83 & 62.51 \\
& SEC      & 69.48  & 63.82      & 67.27 & 87.63  & 66.90 & 82.03 & 55.69 & 63.10 \\
& SSEC     & 70.07  & 65.21      & 67.72 & 87.28  & 67.02 & 81.37 & 56.42 & 63.25 \\
& IPMI     & 73.12  & 65.91      & \underline{68.89} & \underline{88.01}  & \underline{68.25} & \underline{82.78} & \underline{56.47} & \underline{63.91} \\
& PMI (Ours) & \textbf{91.01} & \textbf{75.22} & \textbf{73.10} & \textbf{89.08} & \textbf{69.03} & \textbf{83.20} & \textbf{57.78} & \textbf{64.25} \\
\cmidrule(lr){1-10}	
\multirow{8}{*}{\rotatebox[origin=c]{90}{\texttt{WCKmeans}}} 	 
& None     & 89.80 & 72.38       & 65.66 & \underline{87.84} & 64.85 & 82.17 & 57.22 & 64.12 \\
& ESS      & 90.11 & 73.41       & 64.00 & 87.52  & 65.12 & 81.53 & 56.30 & 63.48 \\
& SESS     & 89.98 & 73.08       & 63.63 & 87.41  & 64.78 & 81.20 & 55.98 & 63.21 \\
& LESS     & 90.11 & 73.33       & 64.56 & 87.00  & 64.91 & 81.05 & 55.62 & 63.07 \\
& SEC      & 88.85 & 70.83       & 63.71 & 86.90  & 64.37 & 80.76 & 56.35 & 63.88 \\
& SSEC     & 89.03 & 70.93       & 63.63 & 87.04  & 64.55 & 80.92 & 56.47 & 63.02 \\
& IPMI     & \underline{90.47} & \underline{74.03} & \underline{66.63} & 87.45 & \underline{66.13} & \underline{82.41} & \underline{57.91} & \underline{64.33} \\
& PMI (Ours) & \textbf{90.65} & \textbf{74.25} & \textbf{72.25} & \textbf{88.60} & \textbf{66.72} & \textbf{83.07} & \textbf{59.28} & \textbf{64.98} \\
\bottomrule
\end{tabular}
}
\end{small}
\vspace{-2ex}
\end{table}

%% file: tex/appendix.tex
\section{Proof of Lemma~\ref{lem:edge-range}}\label{sec:lemma_proof}
\begin{proof}
Let $(t_i,t_j)=\argmax{(t_a,t_b)\in \mathcal{S},\ a<b}{a}$. In the extreme case, $j=i+1$ and only edges $(t_a,t_{a+1}), (t_a,t_{a+2}), \ldots, (t_a,t_{j})$ are added into $\mathcal{S}$ for each $a\in [1,i-1]$ as they are surely larger than $(t_i,t_j)$. Put in another way, for each $t_a$ with $a\in [1,i]$, we can select $t_b$ with $a+1\le b \le j$, i.e., which corresponds to $j-a=i+1-a$ text instances. Thus, the total number of edges satisfy: 
$$1+2+\ldots+i = \frac{(i+1)i}{2} = N.$$
Then, we can derive
\begin{equation*}
i^2+i-2N=0 \Rightarrow i = \frac{\sqrt{8N+1}-1}{2},
\end{equation*}
which finishes the proof.
\end{proof}

\section{Details of Datasets and Baselines}
\stitle{Datasets}
The {\em BBC News}~\cite{keraghel2024beyond} dataset contains 2,225 headlines collected from the BBC News website and categorized into 5 high-level topics. It serves as a benchmark for multi-class classification tasks in the news field. The {\em Tweet}~\cite{yin2016model} dataset consists of 2,472 tweets annotated for relevance to 89 queries from the TREC Microblog track, and is widely used for short-text retrieval and ranking. The {\em Bank77}~\cite{zhang2021discovering} dataset includes 3,080 customer service utterances mapped to 77 intent categories, enabling fine-grained intent classification in the financial category. The {\em Reddit}~\cite{muennighoff2023mteb} dataset, with 3,217 posts from online communities, is used primarily for unsupervised topic discovery. The {\em CLINC}~\cite{zhang2022new} dataset comprises 4,500 user requests labeled across 10 distinct domains for domain-level text clustering. Finally, the {\em Massive}~\cite{fitzgerald2022massive} dataset contains 11,514 utterances spanning 18 scenario-based categories, and is often used in multilingual and multi-scenario classification settings.

\stitle{Baselines}
We include two sets of baselines in the experiments: {\em embedding-based clustering methods} and {\em LLM-assisted clustering methods}.
In embedding-based clustering methods, we include seven representative baselines that follow a two-stage paradigm: generating fixed-dimensional text embeddings followed by conventional clustering algorithms, such as \texttt{K-Means++} or {\em spectral clustering}~\cite{von2007tutorial}.
The selected embedding models encompass both traditional and neural models. 
\begin{itemize}[leftmargin=*]
\item \textbf{\texttt{TF-IDF}}: A classic sparse representation based on term frequency-inverse document frequency, widely used in earlier text clustering studies. \item \textbf{\texttt{E5}}~\cite{wang2022text}: A recent retrieval-oriented embedding model pretrained with a multi-task objective, shown to perform well in clustering tasks. \item \textbf{\texttt{DistilBERT}} and \textbf{\texttt{Sentence-BERT}}~\cite{reimers2019sentence}: Popular lightweight transformers optimized for sentence-level embeddings. 
\item \textbf{\texttt{Instructor-Large}}~\cite{su2023one}: A powerful instruction-tuned model designed to align embeddings with various downstream tasks; we use it as the default embedder in most baselines due to its strong performance. 
\item \textbf{\texttt{OpenAI-GPT}}~\cite{brown2020language} and \textbf{\texttt{LLaMA-2 (7B)}}~\cite{touvron2023llama}: Large language models (LLMs) whose embeddings are obtained by averaging final-layer token representations. As LLMs are often optimized for generation, we treat them as black-box embedding extractors here.
\end{itemize}
These embedding models are coupled with \texttt{K-Means++} or {\em spectral clustering}, depending on the context.

In LLM-assisted clustering methods, we additionally consider three recent strong methods that leverage LLMs beyond embedding extraction. 
\begin{itemize}[leftmargin=*]
\item \textbf{\texttt{SCCL}}~\cite{zhang2021supporting}: A self-supervised contrastive clustering method that improves cluster discrimination by minimizing intra-cluster distance and maximizing inter-cluster margins in embedding space. \item \textbf{\texttt{ClusterLLM}}~\cite{zhang2023clusterllm}: Utilizes LLMs to provide clustering feedback and refines smaller embedding models through LLM guidance. We also include a variant that excludes fine-tuning to assess the impact of feedback alone. 
\item \textbf{\texttt{PO-PCKMeans}}~\cite{viswanathan2023large}: A semi-supervised approach that constructs pairwise constraints from an LLM based on a small set of labeled examples and integrates them into a constrained clustering framework. 
\end{itemize}

\section{Additional Algorithmic Details}\label{sec:additional_algorithms}

\subsection{\texttt{WCKMeans}}

\begin{algorithm}[!t]
\caption{\texttt{WCKMeans}}\label{alg:WCKMeans}
\begin{small}
\KwIn{Corpus $\mathcal{T}$, clusters $K$, MLS $\Pset$, CLS $\Nset$, max iterations $T$}
\KwOut{Labels $\Gamma$}

Normalize $\mathcal{T}$ to get embeddings $\mathbf{X}$;\\
Construct constraint matrix $\mathbf{R}$ with $\Pset$, $\Nset$, $|\mathcal{T}|$, and $\mathbf{X}$;\\
Initialize centers $\mathcal{U} \gets \{\mathbf{u}_1, \ldots, \mathbf{u}_K\}$\;

\While{not converged \& $t < T$}{
    \ForEach{$x_i \in \mathbf{X}$}{

        $\Gamma_i \;\gets\;
        \argmin{1 \le k \le K} 
            \|x_i - \mathbf{u}_k\|^2
            +
            \sum\limits_{\substack{(i,j) \in \Pset \\ \Gamma_j \neq k}} R_{i,j}
            +
            \sum\limits_{\substack{(i,j) \in \Nset \\ \Gamma_j = k}} |R_{i,j}|$

    }
    Update centers $\mathbf{u}_k \gets \texttt{Mean}\bigl(\{\,x_i \mid \Gamma_i = k\}\bigr)$ $\forall{k}$\;
}

\Return{$\Gamma$}
\end{small}
\end{algorithm}

Algorithm \ref{alg:WCKMeans} integrates well-designed weighted constraints into standard \texttt{K-Means} by creating a constraint matrix $\mathbf{R}$, where $R_{i,j}>0$ for must‐links and $R_{i,j}<0$ for cannot‐links. In each iteration, each point $x_i$ is assigned to the cluster $k$ that minimizes its squared distance to centroid $\mathbf{u}_k$ plus penalties: add $R_{i,j}$ if a must‐link partner $j$ is in a different cluster, and add $|R_{i,j}|$ if a cannot‐link partner $j$ is in the same cluster. After assignments, centroids are updated as the mean of their assigned points. This repeats until convergence or $T$ iterations.

\subsection{\texttt{WCSC}}

\begin{algorithm}[!t]
\caption{\texttt{WCSC}}\label{alg:WCSC}
\begin{small}
\KwIn{Corpus $\mathcal{T}$, clusters $K$, MLS $\Pset$, CLS $\Nset$}
\KwOut{Labels $\Gamma$}

Normalize $\mathcal{T}$ to get embeddings $\mathbf{X}$;\\
Build affinity matrix $\mathbf{A}$ with $\mathbf{X}$;\\
Construct constraint matrix $\mathbf{R}$ with $\Pset$, $\Nset$, $|\mathcal{T}|$, and $\mathbf{X}$;\\
$\mathbf{L} \gets \mathbf{D}^{-1/2}(\mathbf{D} - \mathbf{A})\mathbf{D}^{-1/2},\ \mathbf{R}_\alpha \gets \mathbf{D}^{-1/2}\mathbf{R}\mathbf{D}^{-1/2} - \alpha\mathbf{I}$\;
Compute top-$K$ generalized eigenvectors $\mathbf{V}$ of $\mathbf{L}$ and $\mathbf{R}_\alpha$;\\
$\Gamma \gets \texttt{K-Means}(\mathbf{D}^{-1/2}\mathbf{V})$\;

\Return{$\Gamma$}
\end{small}
\end{algorithm}

Algorithm \ref{alg:WCSC} starts by building a graph that captures how similar each pair of points is in the normalized embedding space. Then selecting top-$K$ eigenvectors finds a low‐dimensional embedding that balances the natural data structure with must‐links and cannot‐links. The $\alpha$~\cite{wang2014constrained} serves as a balance factor between the original information captured by the graph and the constraints. Finally, \texttt{K-Means} is applied to produce the cluster labels.

\subsection{Edge Query}

\begin{algorithm}[!t]
\caption{Edge Query}\label{alg:edge_query}
\begin{small}
\KwIn{Corpus $\mathcal{T}$, the number of edges \(N\), Oracle $\mathbf{O}$}
\KwOut{MLS $\Pset$, CLS $\Nset$}

$\mathcal{S} \gets \texttt{GreedyEdgeSelection}(\mathcal{T}, N)$\;
\ForEach{$(t_i, t_j) \in \mathcal{S}$}{
    \Switch{$\mathbf{O}(t_i, t_j)$}{
        \Case{ML}{$\Pset \gets \Pset \cup \{(t_i, t_j)\}$}
        \Case{CL}{$\Nset \gets \Nset \cup \{(t_i, t_j)\}$}
    }
}
\Return{$\Pset$, $\Nset$}
\end{small}
\end{algorithm}

Algorithm \ref{alg:edge_query} uses Algorithm \ref{alg:edge} (\texttt{GreedyEdgeSelection}) to select the top $N$ instance pairs most informative for querying. Each chosen pair $(t_i,t_j)$ is queried to the oracle $\mathbf{O}$, and the corresponding constraints are added based on the responses.

\subsection{Triangle Query}

\begin{algorithm}[!t]
\caption{Triangle Query}\label{alg:triangle_query}
\begin{small}
\KwIn{Corpus $\mathcal{T}$, the number of triangles \(N_\triangle\), Oracle $\mathbf{O}$}
\KwOut{MLS $\Pset$, CLS $\Nset$}

$\mathcal{S}_\triangle \gets \texttt{GreedyTriangleSelection}(\mathcal{T}, N_\triangle)$\;
\ForEach{$(t_i, t_j, t_k) \in \mathcal{S}_\triangle$}{
    \Switch{$\mathbf{O}(t_i, t_j, t_k)$}{
        \Case{all‐same}{\(\forall\,(p,q)\in\{(i,j),(i,k),(j,k)\}:\ \Pset \gets \Pset \cup \{(t_p, t_q)\}\)}
        \Case{ij‐same}{\(\Pset \gets \Pset \cup \{(t_i, t_j)\},\quad \Nset \gets \Nset \cup \{(t_i, t_k),(t_j, t_k)\}\)}
        \Case{ik‐same}{\(\Pset \gets \Pset \cup \{(t_i, t_k)\},\quad \Nset \gets \Nset \cup \{(t_i, t_j),(t_j, t_k)\}\)}
        \Case{jk‐same}{\(\Pset \gets \Pset \cup \{(t_j, t_k)\},\quad \Nset \gets \Nset \cup \{(t_i, t_j),(t_i, t_k)\}\)}
        \Case{all‐diff}{\(\forall\,(p,q)\in\{(i,j),(i,k),(j,k)\}:\ \Nset \gets \Nset \cup \{(t_p, t_q)\}\)}
    }
}
\Return{$\Pset$, $\Nset$}
\end{small}
\end{algorithm}

Algorithm \ref{alg:triangle_query} utilizes Algorithm \ref{alg:triangle} (\texttt{GreedyTriangleSelection}) to select $N_\triangle$ triangles $(t_i,t_j,t_k)$. For each triangle, the oracle $\mathbf{O}$ returns five outcomes—“all‐same”, “ij‐same”, “ik‐same”, “jk-same”, or “all‐diff”, and the corresponding must‐link or cannot‐link constraints are added to $\Pset$ and $\Nset$. 

\input{figs/ablation-embedding}

\section{Additional Experiments}\label{sec:additional_experiments}

\stitle{Various Text Encoders}
Table~\ref{tab:ablation-encoder} reports the clustering performance of \alg{} and strong baselines using two representative encoders: \texttt{Sentence-BERT} and \texttt{Instructor-Large}. Results demonstrate the consistent superiority of \alg{} across datasets and embeddings. For instance, with \texttt{Instructor-Large} on \emph{Tweet}, \alg{} reaches 73.10\% ACC and 89.08\% NMI, exceeding \texttt{ClusterLLM} by 7.29\% and 0.47\%. With \texttt{Sentence-BERT} on \emph{BBC News}, \alg{} achieves 84.45\% ACC and 61.65\% NMI, outperforming \texttt{PO-PCKMeans} by 1.75\% and 4.22\%, respectively. These gains are consistent across diverse corpora, from structured news to informal tweets and task-oriented texts like \emph{Bank77} and \emph{Reddit}. The results highlight the adaptability of \alg{} and its ability to take advantage of powerful pre-trained encoders and informative constraints without task-specific tuning.

\input{figs/ablation-gpt}

\stitle{Queried LLMs}
Fig. \ref{fig:LLMs} illustrates the clustering results of \alg{} when querying four popular LLMs including \texttt{GPT-4o}, \texttt{GPT-4o-Mini}, \texttt{Gemini-1.5-Pro}, and \texttt{Claude-3.5-Sonnet}. Key observations include: (1) Minimal differences in the adoption of various LLMs. (2) LLMs influence answer accuracy, particularly regarding constraints, without affecting query triangle selection; for example, \texttt{GPT-4o} enhances constraint accuracy compared to \texttt{GPT-4o-Mini}, resulting in increased overall accuracy. (3) The advantages of more expensive LLMs do not warrant their costs, underscoring the necessity for more cost-effective models capable of managing complex tasks like triangle queries.

\section{Additional Prompts}

\begin{table*}[!h]
\centering
\caption{Example prompts for edge and triangle queries on different datasets.}
\vspace{-2ex}
\label{tab:more-prompts}
\renewcommand{\arraystretch}{0.93}
\resizebox{\textwidth}{!}{
\begin{tabular}{c|c|l}
\hline
\textbf{Dataset} & \textbf{Type} & \textbf{Prompt} \\
\hline
\multirow{2}{*}{BBC News} 
  & Edge Query & Cluster BBC News docs by whether they belong to the same news category. For each pair, respond with Yes or No without explanation. \\
  &            & - News \#1: Ad sales boost Time Warner profit \\
  &            & - News \#2: Air passengers win new EU rights \\
  &            & Given this context, do News \#1 and News \#2 likely correspond to the same news category? \\
  & Triangle Query & Cluster BBC News docs by whether they belong to the same news category. For each triangle, respond with a, b, c, d, or e without explanation. \\
  &            & - News \#1: REM announce new Glasgow concert \\
  &            & - News \#2: Moreno debut makes Oscar mark \\
  &            & - News \#3: Last Star Wars 'not for children' \\
  &            & Given this context, do News \#1, News \#2, and News \#3 likely correspond to the same news category? \\
  &            & \textbf{(a)} All are same category.\quad \textbf{(b)} Only \#1 and \#2 are same category.\quad \textbf{(c)} Only \#1 and \#3 are same category. \\
  &            & \textbf{(d)} Only \#2 and \#3 are same category.\quad \textbf{(e)} None. \\
\hline
\multirow{2}{*}{Tweet}
  & Edge Query & Cluster Tweet docs by whether they belong to the same tweet category. For each pair, respond with Yes or No without explanation. \\
  &            & - Tweet \#1: super bowl commercial \\
  &            & - Tweet \#2: kung pao chicken recipe \\
  &            & Given this context, do Tweet \#1 and Tweet \#2 likely correspond to the same tweet category? \\
  & Triangle Query & Cluster Tweet docs by whether they belong to the same tweet category. For each triangle, respond with a, b, c, d, or e without explanation. \\
  &            & - Tweet \#1: weight loss diet plan acai juice equal nutritional claim acai \\
  &            & - Tweet \#2: watch christina aguilera screw national anthem super bowl post \\
  &            & - Tweet \#3: yell cast jr hartley ad digital era author search fly fishing book dj hunt book read \\
  &            & Given this context, do Tweet \#1, Tweet \#2, and Tweet \#3 likely correspond to the same tweet category? \\
  &            & \textbf{(a)} All are same category.\quad \textbf{(b)} Only \#1 and \#2 are same category.\quad \textbf{(c)} Only \#1 and \#3 are same category. \\
  &            & \textbf{(d)} Only \#2 and \#3 are same category.\quad \textbf{(e)} None. \\
\hline
\multirow{2}{*}{Bank77}
  & Edge Query & Cluster Bank77 docs by whether they belong to the same intent category. For each pair, respond with Yes or No without explanation. \\
  &            & - Intent \#1: I think something went wrong with my card delivery as I haven't received it yet. \\
  &            & - Intent \#2: How do I link to my credit card with you? \\
  &            & Given this context, do Intent \#1 and Intent \#2 likely correspond to the same intent category? \\
  & Triangle Query & Cluster Bank77 docs by whether they belong to the same intent category. For each triangle, respond with a, b, c, d, or e without explanation. \\
  &            & - Intent \#1: What is the current exchange rate for me? \\
  &            & - Intent \#2: I believe my card payment exchange rate is incorrect. \\
  &            & - Intent \#3: I think my child used my card while I wasn't home. \\
  &            & Given this context, do Intent \#1, Intent \#2, and Intent \#3 likely correspond to the same intent category? \\
  &            & \textbf{(a)} All are same category.\quad \textbf{(b)} Only \#1 and \#2 are same category.\quad \textbf{(c)} Only \#1 and \#3 are same category. \\
  &            & \textbf{(d)} Only \#2 and \#3 are same category.\quad \textbf{(e)} None. \\
\hline
\multirow{2}{*}{Reddit}
  & Edge Query & Cluster Reddit docs by whether they belong to the same topic category. For each pair, respond with Yes or No without explanation. \\
  &            & - Topic \#1: CTA to make all rail stations accessible within 20 years \\
  &            & - Topic \#2: A Krazy Mug Oyee Balle Balle!!! Kettle with set of two glasses \\
  &            & Given this context, do Topic \#1 and Topic \#2 likely correspond to the same topic category? \\
  & Triangle Query & Cluster Reddit docs by whether they belong to the same topic category. For each triangle, respond with a, b, c, d, or e without explanation. \\
  &            & - Topic \#1: Batman and Batwoman Fan art \\
  &            & - Topic \#2: Books on American Union and Labour Movements? \\
  &            & - Topic \#3: When Taliban offer you gold: Afghan youth in crisis? \\
  &            & Given this context, do Topic \#1, Topic \#2, and Topic \#3 likely correspond to the same topic category? \\
  &            & \textbf{(a)} All are same category.\quad \textbf{(b)} Only \#1 and \#2 are same category.\quad \textbf{(c)} Only \#1 and \#3 are same category. \\
  &            & \textbf{(d)} Only \#2 and \#3 are same category.\quad \textbf{(e)} None. \\
\hline
\multirow{2}{*}{Massive}
  & Edge Query & Cluster Massive docs by whether they belong to the same scenario category. For each pair, respond with Yes or No without explanation. \\
  &            & - Scenario \#1: here is something from today \\
  &            & - Scenario \#2: remind me of the meeting on tuesday \\
  &            & Given this context, do Scenario \#1 and Scenario \#2 likely correspond to the same scenario category? \\
  & Triangle Query & Cluster Massive docs by whether they belong to the same scenario category. For each triangle, respond with a, b, c, d, or e without explanation. \\
  &            & - Scenario \#1: alexa play my country playlist \\
  &            & - Scenario \#2: write email to my company mate to submit the task tomorrow \\
  &            & - Scenario \#3: remind me about my business meeting at three and forty five p. m. \\
  &            & Given this context, do Scenario \#1, Scenario \#2, and Scenario \#3 likely correspond to the same scenario category? \\
  &            & \textbf{(a)} All are same category.\quad \textbf{(b)} Only \#1 and \#2 are same category.\quad \textbf{(c)} Only \#1 and \#3 are same category. \\
  &            & \textbf{(d)} Only \#2 and \#3 are same category.\quad \textbf{(e)} None. \\
\hline
\end{tabular}
}
\end{table*}

We provide additional example prompts in Table \ref{tab:more-prompts} for edge and triangle queries across multiple datasets, including {\em BBC News}, {\em Tweet}, {\em Bank77}, {\em Reddit}, and {\em Massive}. Edge queries ask whether two documents belong to the same category (Yes/No), while triangle queries assess the categorical relationship among three documents via a five-way multiple-choice format. These prompts are designed to support constraint construction using LLMs for clustering tasks.

%% file: figs/ablation-embedding.tex
\begin{table*}[!t]
\centering
\renewcommand{\arraystretch}{0.8}
\caption{Clustering performance with various text encoders}
\label{tab:ablation-encoder}
\vspace{-2ex}
\begin{tabular}{c|c|cc|cc|cc|cc}
\toprule
\multirow{2}{*}{\textbf{Encoder}} 
& \multirow{2}{*}{\textbf{Method}} 
& \multicolumn{2}{c|}{\textbf{BBC News}} 
& \multicolumn{2}{c|}{\textbf{Tweet}} 
& \multicolumn{2}{c|}{\textbf{Bank77}} 
& \multicolumn{2}{c}{\textbf{Reddit}} \\
\cline{3-10}
 &  & \textbf{ACC} & \textbf{NMI} 
 & \textbf{ACC} & \textbf{NMI} 
 & \textbf{ACC} & \textbf{NMI} 
 & \textbf{ACC} & \textbf{NMI} \\
\midrule
\multirow{4}{*}{\texttt{Sentence-BERT}} 
& \texttt{SCCL}         & 82.00 & \underline{58.33} & 53.60 & 72.80 & 27.80 & 53.21 & 36.60 & 45.71 \\
& \texttt{ClusterLLM}   & 80.08 & 54.13             & 57.47 & 83.74 & \underline{58.61} & 77.06 & 36.09 & 44.74 \\
& \texttt{PO-PCKMeans}  & \underline{82.70} & 57.43 & \underline{61.39} & \underline{86.12} & 57.73 & \underline{78.09} & \underline{38.48} & \underline{48.56} \\
& \alg{}                & \textbf{84.45} & \textbf{61.65} & \textbf{78.48} & \textbf{89.30} & \textbf{67.05}  & \textbf{80.46}  & \textbf{40.94}  & \textbf{49.32} \\
\hline
\multirow{4}{*}{\texttt{Instructor-Large}} 
& \texttt{SCCL}         & 83.60 & 62.00             & 36.50 & 68.50 & 36.00 & 61.30 & 28.70 & 36.00 \\
& \texttt{ClusterLLM}   & 88.40 & 69.73             & \underline{65.81} & \underline{88.61} & \underline{68.27} & \underline{82.45} & 54.80 & 62.47 \\
& \texttt{PO-PCKMeans}  & \underline{89.53} & \underline{72.16} & 65.58 & 88.20 & 67.79 & 81.92 & \underline{55.10} & \underline{62.51}  \\
& \alg{}                & \textbf{91.01} & \textbf{75.22} & \textbf{73.10} & \textbf{89.08} &\textbf{69.03}& \textbf{83.20}& \textbf{57.78}& \textbf{64.25}  \\
\bottomrule
\end{tabular}
\vspace{1ex}
\end{table*}

%% file: figs/ablation-gpt.tex
\begin{figure}[!t]
\centering
\begin{small}
\begin{tikzpicture}
\begin{customlegend}[
        legend entries={\texttt{GPT-4o},\texttt{GPT-4o-Mini}},
        legend columns=4,
        area legend,
        legend style={at={(0.45,1.15)},anchor=north,draw=none,font=\small,column sep=0.25cm}]
        \addlegendimage{preaction={fill, lightblue}, pattern={grid}}  
        \addlegendimage{preaction={fill, myorange}, pattern={crosshatch dots}}    
    \end{customlegend}
\end{tikzpicture}
\begin{tikzpicture}
\begin{customlegend}[
        legend entries={\texttt{Gemini-1.5-Pro},\texttt{Claude-3.5-Sonnet}},
        legend columns=4,
        area legend,
        legend style={at={(0.45,1.15)},anchor=north,draw=none,font=\small,column sep=0.25cm}]
        \addlegendimage{preaction={fill, mypink},pattern=north west lines} 
        \addlegendimage{preaction={fill, mycyan},pattern=crosshatch}    
    \end{customlegend}
\end{tikzpicture}
\\[-\lineskip]
\vspace{-4mm}
\subfloat[{\em BBC News}]{
\begin{tikzpicture}[scale=1]
\begin{axis}[
    height=\columnwidth/2.5,
    width=\columnwidth/2.0,
    xtick=\empty,
    ybar=1.0pt,
    bar width=0.4cm,
    enlarge x limits=true,
    ylabel={\em Acc},
    xticklabel=\empty,
    ymin=0.84,
    ymax=0.92,
    ytick={0.84,0.86,0.88,0.9,0.92},
    yticklabels={0.84,0.86,0.88,0.9,0.92},
    xticklabel style = {font=\small},
    yticklabel style = {font=\small},
    every axis y label/.style={at={(current axis.north west)},right=2mm,above=0mm},
    legend style={draw=none, at={(1.02,1.02)},anchor=north west,cells={anchor=west},font=\small},
    legend image code/.code={ \draw [#1] (0cm,-0.1cm) rectangle (0.3cm,0.15cm); },
    ]

\addplot [preaction={fill, lightblue}, pattern={grid}] coordinates {(1,0.9135) }; 
\addplot [preaction={fill, myorange}, pattern={crosshatch dots}] coordinates {(1,0.9101) }; 
\addplot [preaction={fill, mypink},pattern=north west lines] coordinates {(1,0.9121) }; 
\addplot [preaction={fill, mycyan},pattern=crosshatch] coordinates {(1,0.9142) };

\end{axis}
\end{tikzpicture}\hspace{4mm}\label{fig:time-photos}%
}%
\subfloat[{\em Tweet}]{
\begin{tikzpicture}[scale=1]
\begin{axis}[
    height=\columnwidth/2.5,
    width=\columnwidth/2.0,
    xtick=\empty,
    ybar=1.0pt,
    bar width=0.4cm,
    enlarge x limits=true,
    ylabel={\em Acc},
    xticklabel=\empty,
    ymin=0.70,
    ymax=0.74,
    ytick={0.7,0.71,0.72,0.73,0.74},
    yticklabels={0.7,0.71,0.72,0.73,0.74},
    xticklabel style = {font=\small},
    yticklabel style = {font=\small},
    every axis y label/.style={at={(current axis.north west)},right=2mm,above=0mm},
    legend style={draw=none, at={(1.02,1.02)},anchor=north west,cells={anchor=west},font=\small},
    legend image code/.code={ \draw [#1] (0cm,-0.1cm) rectangle (0.3cm,0.15cm); },
    ]

\addplot [preaction={fill, lightblue}, pattern={grid}] coordinates {(1,0.7333) }; 
\addplot [preaction={fill, myorange}, pattern={crosshatch dots}] coordinates {(1,0.7310) }; 
\addplot [preaction={fill, mypink},pattern=north west lines] coordinates {(1,0.7342) }; 
\addplot [preaction={fill, mycyan},pattern=crosshatch] coordinates {(1,0.7327) };

\end{axis}
\end{tikzpicture}\hspace{0mm}\label{fig:time-Cora}%
}%
\end{small}
\vspace{-3ex}
\caption{The performance of \alg{} with various LLMs} \label{fig:LLMs}
\vspace{-1ex}
\end{figure}
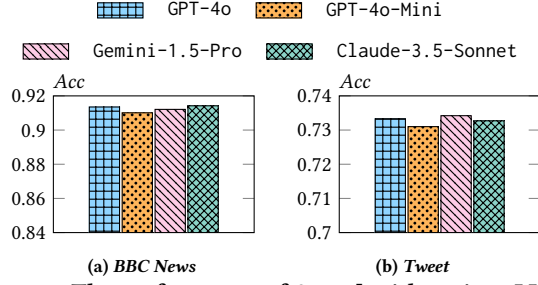